\documentclass[conference]{IEEEtran}
\IEEEoverridecommandlockouts
\usepackage{cite}        
\usepackage{amsmath,amssymb,amsfonts}
\usepackage{algorithmic}
\usepackage{graphicx}
\usepackage{textcomp}
\usepackage{xcolor}
\usepackage{amsmath}     
\usepackage{amssymb}     
\usepackage{amsthm}      
\usepackage{bm}          
\usepackage{mathtools}   
\usepackage{booktabs}    
\usepackage{multirow}    
\usepackage{subcaption}
\usepackage{float}        
\newtheorem{proposition}{Proposition}
\usepackage[switch]{lineno}

\newcommand{\Cat}{\mathrm{Cat}}  

\def\BibTeX{{\rm B\kern-.05em{\sc i\kern-.025em b}\kern-.08em
    T\kern-.1667em\lower.7ex\hbox{E}\kern-.125emX}}
\usepackage{xcolor}    
\usepackage[colorlinks=true, citecolor=blue]{hyperref}

\begin{document}

\title{Deep Generative Methods and Tire Architecture Design}
\author{
\IEEEauthorblockN{Fouad Oubari}
\IEEEauthorblockA{
\textit{ENS Paris-Saclay, Centre Borelli}\\
\textit{Manufacture Française des Pneumatiques Michelin} \\
Gif-sur-Yvette, France \\
fouad.oubari@ens-paris-saclay.fr}
\and
\IEEEauthorblockN{Raphael Meunier}
\IEEEauthorblockA{
\textit{Manufacture Française des Pneumatiques Michelin} \\
Clermont-Ferrand, France \\
raphael.meunier@michelin.com}
\and
\IEEEauthorblockN{Rodrigue Décatoire}
\IEEEauthorblockA{
\textit{Manufacture Française des Pneumatiques Michelin} \\
Clermont-Ferrand, France \\
rodrigue.decatoire@michelin.com}
\and
\IEEEauthorblockN{Mathilde Mougeot}
\IEEEauthorblockA{
\textit{Université Paris-Saclay, CNRS, ENS Paris-Saclay, Centre Borelli} \\
Gif-sur-Yvette, France \\
mathilde.mougeot@ens-paris-saclay.fr \\
\textit{ENSIIE}\\
Évry, France}
}

\maketitle

\begin{abstract}
As deep generative models proliferate across the AI landscape, industrial practitioners still face critical yet unanswered questions about which deep generative models best suit complex manufacturing design tasks.
This work addresses this question through a complete study of five representative models (Variational Autoencoder, Generative Adversarial Network, multimodal Variational Autoencoder, Denoising Diffusion Probabilistic Model, and Multinomial Diffusion Model) on industrial tire architecture generation.
Our evaluation spans three key industrial scenarios: (i) unconditional generation of complete multi-component designs, (ii) component-conditioned generation (reconstructing architectures from partial observations), and (iii) dimension-constrained generation (creating designs that satisfy specific dimensional requirements). To enable discrete diffusion models to handle conditional scenarios, we introduce categorical inpainting, a mask-aware reverse diffusion process that preserves known labels without requiring additional training. Our evaluation employs geometry-aware metrics specifically calibrated for industrial requirements, quantifying spatial coherence, component interaction, structural connectivity, and perceptual fidelity. Our findings reveal that diffusion models achieve the strongest overall performance; a masking-trained VAE nonetheless outperforms the multimodal variant MMVAE\textsuperscript{+} on nearly all component-conditioned metrics, and within the diffusion family MDM leads in-distribution whereas DDPM generalises better to out-of-distribution dimensional constraints.
\footnote{An anonymized implementation is available at 
\url{https://anonymous.4open.science/r/DGMTA-5122/}. 
A fully documented, open‑source repository will be released upon acceptance.}

\end{abstract}

\section{Introduction}

The design of complex industrial products, such as pneumatic tires, traditionally involves costly iterations of finite-element simulation and physical prototyping. As design complexity increases, reducing these iterative cycles through computational approaches represents significant industrial value. Deep generative models (DGMs) offer a promising direction, providing trainable, differentiable frameworks for modeling complex geometric objects with potential applications in design exploration and optimization.

While Variational Auto-Encoders (VAEs) \cite{kingma2013auto}, Generative Adversarial Networks (GANs) \cite{goodfellow2014generative}, and diffusion models \cite{ho2020denoising} have demonstrated strong performance on canonical image datasets, systematic evaluations of their effectiveness on real industrial data remain limited. Most prior work evaluates a single model class or relies on synthetic datasets like CIFAR-10 or ShapeNet \cite{chang2015shapenet}, leaving open questions about how these models perform on complex multi-component industrial designs with specific structural constraints.

In this work, we perform a controlled methodological comparison of five representative DGMs, VAE, GAN, MMVAE$^{+}$~\cite{palumbo2023mmvae+}, Denoising Diffusion Probabilistic Model (DDPM)~\cite{ho2020denoising}, and Multinomial Diffusion Model (MDM)~\cite{hoogeboom2021argmax},  using a proprietary tire-architecture dataset as an internal benchmark. Importantly, the insights obtained from our evaluation are methodological and thus transferable, independent of the particular dataset employed. We investigate model capabilities across three key industrial scenarios:

\begin{enumerate}
    \item \textbf{Unconditional generation}: Generating complete tire architectures without prior constraints, evaluating models' ability to capture the joint distribution of multiple interacting components.
    
    \item \textbf{Component-conditioned generation}: Reconstructing complete architectures from partial observations of one component, simulating scenarios where designers iterate on specific elements while maintaining overall structural coherence.
    
    \item \textbf{Dimension-constrained generation}: Creating tire architectures that satisfy specific dimensional requirements, addressing the critical industrial need for designs that fit within manufacturing molds and meet industrial specifications.
\end{enumerate}

For MDM specifically, we implement categorical inpainting by adapting the reverse diffusion process with a masked reverse step, enabling conditional generation while maintaining component separation. This adaptation is particularly valuable for industrial applications where component boundaries must remain distinct.

Our contributions include: (1) a comprehensive benchmark of five DGM families under identical parameter budgets on tire architecture generation tasks; (2) a novel implementation of categorical inpainting for multinomial diffusion; and (3) an industrial-relevant evaluation protocol using metrics for spatial coherence, component interaction, structural connectivity, and perceptual fidelity \cite{heusel2017gans}.

\section{Related Work}

\subsection{From Classical Optimization to Deep Generative Models}

Traditional industrial design methods involve expensive iterative physical tests and complex numerical simulations,
 often severely limiting exploration of the design space. For example, automotive wind-tunnel testing incurs 
 substantial daily costs, rapidly escalating project budgets \cite{ashton2020towards}.
 Additionally, critical performance metrics (e.g., footprint shape, vertical stiffness or rolling resistance in tire modeling) are usually assessed with computationally expensive, non-differentiable finite-element analyses (FEA), further constraining practical optimization \cite{fathi2024modeling}.

Automated classical methods such as genetic algorithms and surrogate-assisted optimization like Efficient Global optimization have partially alleviated these challenges by reducing manual effort and simulation costs \cite{obayashi1997comparison,jones1998efficient}. However, these techniques remain fundamentally constrained by high dimensionality and the absence of gradient-based optimization methods \cite{eiben2015introduction}.

The inherent limitations of these classical approaches have motivated the shift toward deep generative models, offering richer representational capacity and differentiable structures for efficient design exploration.

\subsection{Industrial Applications of VAEs and GANs}

Early deep generative models such as VAEs \cite{kingma2013auto} and GANs \cite{goodfellow2014generative} demonstrated significant potential for industrial design optimization by providing differentiable generative priors. Notable industrial applications include airfoil inverse design using conditional GANs \cite{achour2020development}, airfoil parameterization with conditional VAEs for multi-objective optimization \cite{yonekura2022generating}, anechoic coating optimization with VAEs for rapid topology generation \cite{sun2022variational}, and sparse material design where Binded-VAEs outperformed traditional methods in generating realistic industrial rubber compounds \cite{oubari2021binded}.

Despite these successes, VAEs and GANs exhibit inherent limitations when applied to complex industrial data. GANs often suffer from mode collapse and training instability, whereas VAEs typically produce samples with limited geometric precision \cite{10.1007/978-3-031-62281-6_17}. Furthermore, both approaches frequently struggle with modeling inter-component relations in multi-part systems, which is critical for industrial applications like tire architecture design.

\subsection{Multimodal and Structured Generative Models}

Multimodal deep generative models \cite{suzuki2022survey} address the limitations of standard generative approaches by integrating heterogeneous data streams, enhancing generative capabilities through complementary modality information. This capability is particularly valuable for industrial tasks requiring coherent cross-modal representations, modality-specific precision, and robust conditional generation.

The evolution of multimodal VAEs has progressed through several key advancements, including MVAE \cite{wu2018multimodal}, MMVAE \cite{shi2019variational}, MoPoE-VAE \cite{sutter2020multimodal}, Generalised-ELBO MMVAE \cite{sutter2021generalized}, and most recently MMVAE+ \cite{palumbo2023mmvae+}. MMVAE+ specifically addresses the challenge of balancing coherence and modality-specific precision, making it a suitable multimodal baseline for our comparative study.

Recent application-driven studies demonstrate how multimodal DGMs can directly support industrial optimization pipelines. Meta-VAE introduces a two-level architecture in which a Meta-Generator outputs latent codes for several pretrained marginal VAEs, letting the model coordinate multiple unitary components (e.g., two nested cylinders) so that the assembled system respects global contact and equilibrium constraints \cite{10.1007/978-3-031-62281-6_17}.
At the micro-structural scale, Sardeshmukh \textit{et al.}\ fuse a VAE with a regression head and a Gaussian-mixture conditional prior; the resulting shared latent space enables direct one-to-many inverse design of 3-D microstructures that meet target elastic properties without an external optimization loop \cite{sardeshmukh2024material}. 

Going a step further, the PSP-GEN framework, a deep generative probabilistic model of the full process–structure–property chain, embeds process parameters, microstructure images, and properties into two coordinated latent codes (one deterministic and one stochastic), enabling gradient-based inversion for manufacturable two-phase materials at prescribed permeability levels \cite{zang2025psp}.

These application-focused multimodal models illustrate how latent generative spaces can be used for industrial design.


\subsection{Diffusion Models in Industrial Design}

Diffusion models have recently emerged as a powerful alternative to traditional generative approaches, combining the stability of VAEs, the high fidelity of GANs, and the explicit likelihood evaluation capabilities missing in other frameworks. The foundational Denoising Diffusion Probabilistic Model (DDPM) \cite{ho2020denoising} and score-based SDE methods \cite{song2020score} have demonstrated state-of-the-art performance on standard image generation benchmarks.

An important extension for industrial applications is multinomial diffusion \cite{hoogeboom2021argmax}, which adapts the diffusion framework to discrete data such as categorical segmentation maps. 
This approach naturally handles data in which each pixel belongs to exactly one component or background class, thereby enforcing mutual exclusivity between components.
Recent industrial applications of diffusion models include TopoDiff for structural design optimization \cite{maze2023diffusion} and integrated diffusion-CNN approaches for microstructure design of multifunctional composites \cite{lee2024data}. These successful applications provide strong motivation for adopting and extending diffusion methods for our structured component generation tasks.

\subsection{Benchmarking Generative Models for Industrial Applications}

To the best of our knowledge, systematic benchmarks that compare multiple DGMs under identical conditions are still rare in the industrial‐design literature, yet a few recent works offer useful precedents. In materials science, \cite{yue2025benchmarking} evaluated six generative architectures (VAEs, Adversarial Autoencoders, several GAN variants, and an RNN baseline), on a common polymer dataset, reporting validity, novelty, and property‐matching scores. Mechanical‐metamaterial research has likewise seen a four-way comparison (VAE, GAN, WGAN, DDPM) on kirigami cut-pattern generation, which highlighted the difficulty of satisfying hard geometric constraints with Euclidean latent spaces \cite{felsch2024generative}. For large disordered materials, the \textsc{Dismai-Bench} framework contrasted graph-based and coordinate-based diffusion models (plus point-cloud GANs), demonstrating the importance of symmetry-aware data representations \cite{yong2024dismai}. Outside design, vibration-based fault diagnosis in rotating machinery has been benchmarked with convolutional and variational autoencoders, revealing trade-offs between reconstruction fidelity and anomaly sensitivity \cite{radicioni2025vibration}. Collectively, these works establish that no single family of DGMs dominates across domains and metrics.

These benchmarks focus on single-object or graph-level structures and evaluate unconditional generation in isolation, our study offers a complementary perspective by tackling a realistic, multi-component industrial system.

\section{Methods}
      
\begin{figure}[H]
  
  \centering

  \includegraphics[width=\linewidth]{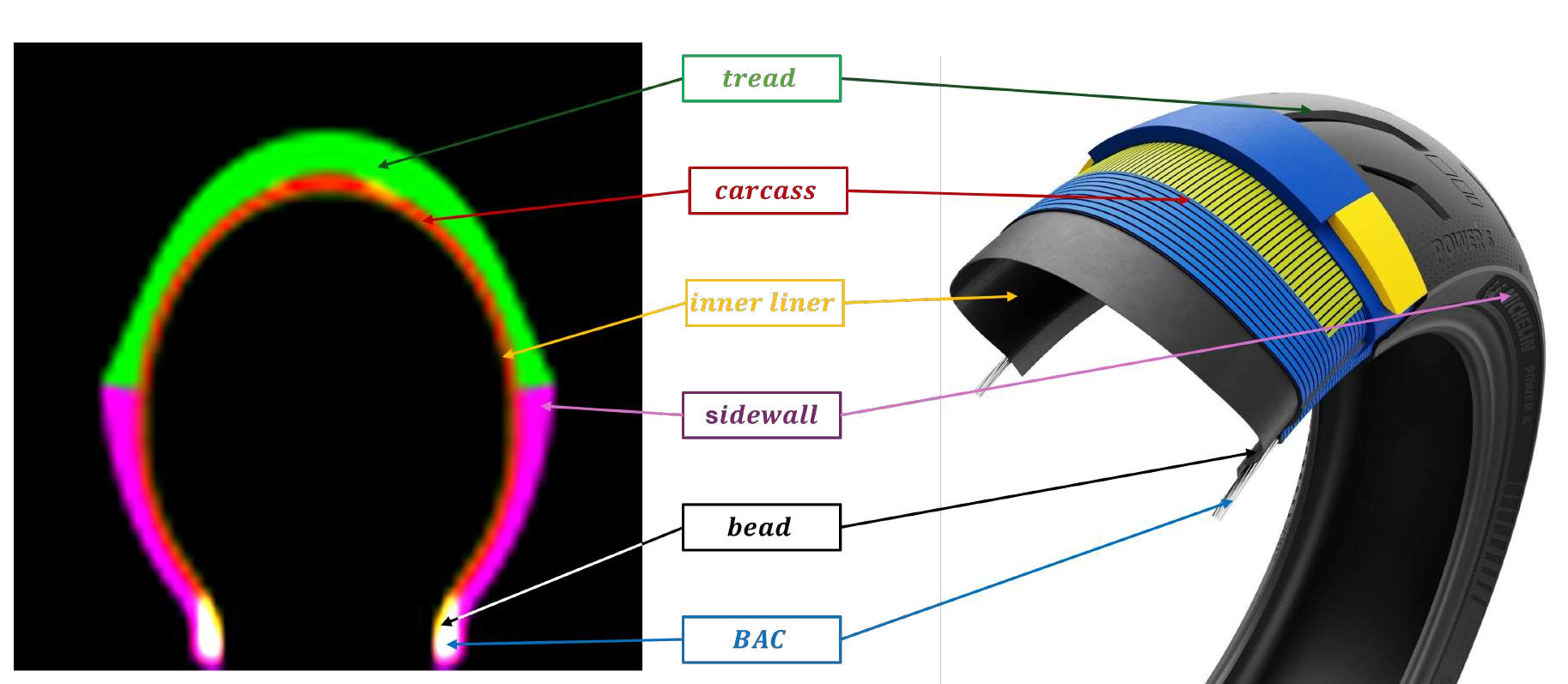}  \caption{Tire architecture representation used in this work. \textbf{Left:} Visualization of the orthogonal cross-section showing six key components: (\texttt{tread}) provides tread contact with the road and structural rigidity; (\texttt{carcass}) consists of textile plies extending from bead to bead forming the main structural framework; (\texttt{inner liner}) is a thin, airtight rubber layer that seals the internal cavity; outer (\texttt{sidewall}) protects the carcass and absorbs deformation during operation; (\texttt{bead filler}) comprises metallic rings that secure the tire to the rim and ensure airtight sealing when mounted; and (\texttt{bead-apex cushion}- (\texttt{BAC})) reinforces the sidewall above the bead. \textbf{Right:} Photorealistic model illustrating the physical placement of these components in a tire architecture. Note that our representation shows an orthogonal cross-section, not the full three-dimensional tire.}
\label{fig:tire_representation}
\end{figure}

The following section outlines the methodological framework for evaluating different deep generative models on tire architecture data. We begin by establishing consistent notation, then describe each generative model and its adaptation for both unconditional and conditional generation tasks.

\subsection{Notation}
\label{sec:methods:notation}

We establish the following notation conventions to ensure consistency throughout the paper:

\begin{itemize}
    \item $\mathbf{X}$ denotes a complete tire architecture, defined as $\mathbf{X} = (\mathbf{x}_1, \ldots, \mathbf{x}_M)$, where $M$ is the number of components (in our dataset, $M=6$).
    
    \item $\mathbf{x}_i \in \mathbb{R}^{H \times W}$ represents the $i$-th component as a grayscale image, where $H=W=64$ in our implementation.
    
    \item $\bar{\mathbf{x}} \in \mathbb{R}^{M \times H \times W}$ denotes the stacked representation of all components, used as input for most models.
    
    \item Categorical representation (MDM): for each pixel
        $i\in\{1,\dots,H\!\times\!W\}$ we define
        $\mathbf c_i\in\{0,1\}^{K}$ as its one-hot label
        ($K=M{+}1$, i.e.\ $M$ components plus background).
        For simplicity we write $\mathbf c$ instead of $\mathbf c_i$
whenever the index is unambiguous.

    \item $\mathbf{z} \in \mathbb{R}^d$ represents the shared latent variable in all generative models, where $d$ is the dimension of the latent space.
    
    \item For MMVAE+, $\mathbf{w}_i \in \mathbb{R}^{d_i}$ denotes the component-specific latent variable for the $i$-th component, and $\mathbf{W} = (\mathbf{w}_1, \ldots, \mathbf{w}_M)$ refers to the collection of all component-specific latent variables.
    
    \item For diffusion models, $\mathbf{x}_t$ denotes the noisy version of the data at timestep $t$ in the forward diffusion process. For categorical diffusion models, $\mathbf{c}_{t,i}$ represents the probability vector for pixel $i$ at timestep $t$.
    

    \item $\beta_t$ is the noising ­rate at step $t$:  
      for DDPM it is the Gaussian variance coefficient \cite{ho2020denoising},  
      whereas for MDM it is the categorical swap probability \cite{hoogeboom2021argmax}, and $\alpha_t := 1-\beta_t$

      \item $\mathbf{d} = (w, h)$ represents the dimensional parameters of a tire architecture, where $w$ is the width and $h$ is the height.
    
    \item $W_1$ denotes the 1-Wasserstein distance used to compute various evaluation metrics.
    
    \item For evaluation metrics, we use IoU for Intersection-over-Union, CoM for Center of Mass, RCE for Region-Connectivity Error, FID for Fréchet Inception Distance, and DimErr for Dimension Error.
    
    \item For classifier-free guidance, $w$ denotes the guidance scale controlling the influence of conditional information during the generation process.

  \end{itemize}


\subsection{Dataset and Model-Specific Preprocessing}

Our proprietary dataset consists of 1,130 tire architectures, each composed of six distinct structural components: $\{\mathit{bead\ filler}, \mathit{cushion},  \mathit{carcass}, \mathit{sidewall}, \mathit{tread}, \mathit{inner\ liner}\}$ (Figure \ref{fig:tire_representation}). More industrial details about these components and their functional roles can be found in the appendix. Each component is represented as a grayscale image of resolution $64 \times 64$. We partition the dataset into 80\% for training and 20\% for testing.

Different model architectures require specific input formats:

\begin{enumerate}
    \item \textbf{VAE, GAN, DDPM:} We stack the six component images into a single multi-channel tensor $\bar{\mathbf{x}} \in \mathbb{R}^{6 \times 64 \times 64}$.
    
    \item \textbf{MMVAE+:} Each component $\mathbf{x}_i$ is treated as a separate modality and processed by a dedicated encoder.
    
    \item \textbf{MDM:} Following the one-hot convention introduced in
      Section \ref{sec:methods:notation} (Notation),
      we convert the 6-channel stack into a
      $6$-channel one-hot tensor
      $\mathbf c\in\{0,1\}^{6\times64\times64}$, so exactly
      one channel is active per pixel (background or one component).

\end{enumerate}

\begin{figure}[t]
  \centering
  \includegraphics[width=.7\linewidth]{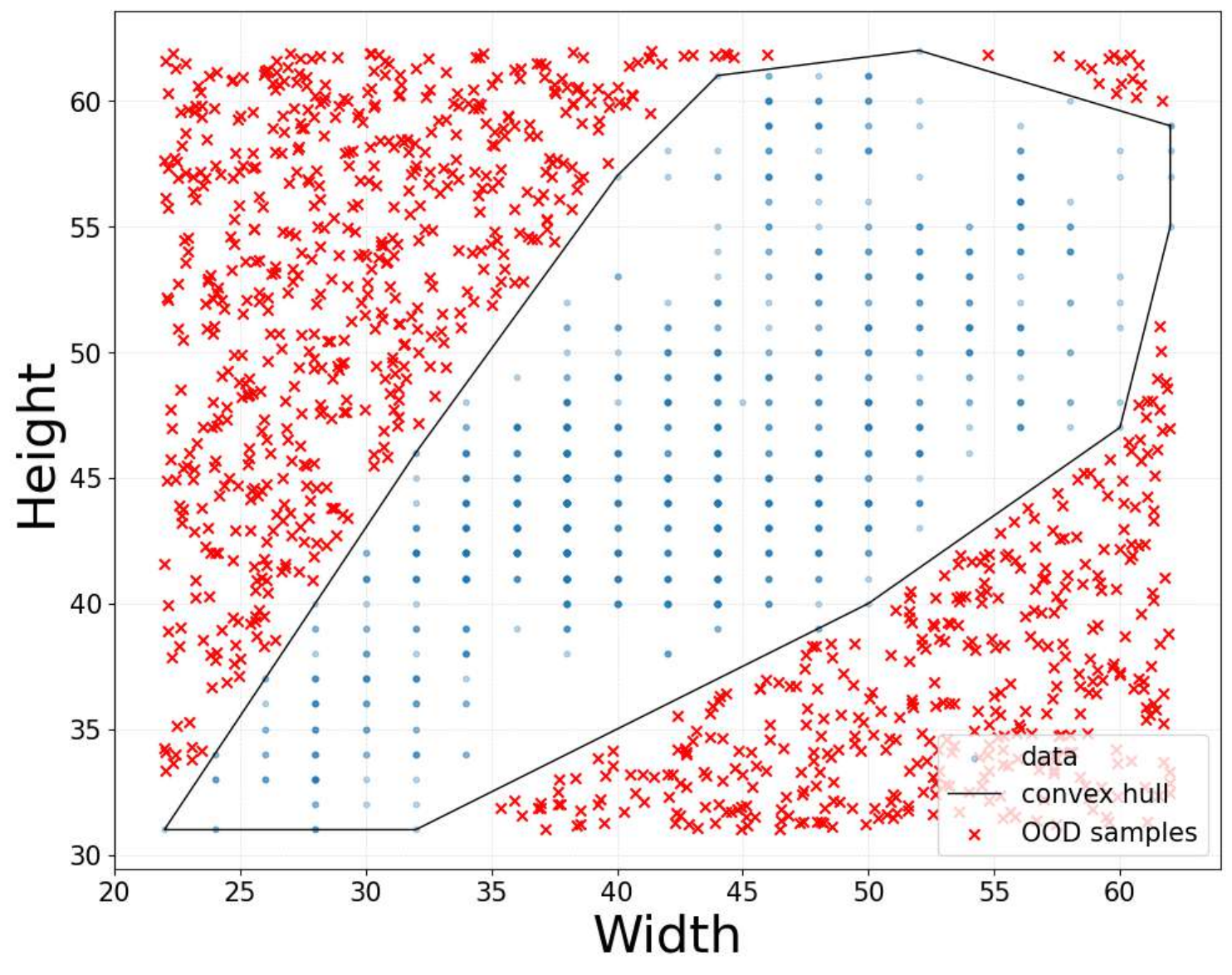}
  \caption{Distribution of tire dimensional attributes (height and width) in our dataset. The convex hull (outlined) defines the boundary between in-distribution (ID) and out-of-distribution (OOD) specifications. While ID points represent common dimensional configurations, OOD points test model generalization to novel but physically plausible dimensional constraints.}
  \label{fig:id_ood_hull}
\end{figure}

For dimension-constrained generation, each tire architecture in our dataset is additionally annotated with its height and width measurements. Figure \ref{fig:id_ood_hull} illustrates the distribution of these dimensional attributes across our dataset. We compute the convex hull of this distribution to define the boundary between in-distribution (ID) and out-of-distribution (OOD) dimensional specifications. This distinction enables us to evaluate model generalization not only on test samples with dimensions similar to the training data (ID test set) but also on more challenging dimensional specifications that lie outside the training distribution (OOD test set).

The OOD test specifications are sampled to maintain physical plausibility while exploring dimensional extremes. For example, we avoid unrealistic specifications (such as zero-height tires) but include challenging cases with exaggerated aspect ratios or overall dimensions that might be required for novel vehicle designs.

\subsection{Generative Models}

We implement and evaluate six representative deep generative models:

\subsubsection{Generative Adversarial Network (GAN)}

The GAN architecture \cite{goodfellow2014generative} involves a competitive training process between a generator $G_\theta$ and a discriminator $D_\psi$. The generator maps latent samples drawn from a Gaussian prior $\mathbf{z} \sim \mathcal{N}(\mathbf{0}, \mathbf{I})$ to multi-channel images $\bar{\mathbf{x}} \in \mathbb{R}^{M \times H \times W}$. The GAN objective optimizes the minimax formulation:
\begin{equation}
\min_\theta \max_\psi \mathbb{E}_{\bar{\mathbf{x}} \sim p_{\text{data}}}[\log D_\psi(\bar{\mathbf{x}})] + \mathbb{E}_{\mathbf{z} \sim \mathcal{N}(\mathbf{0}, \mathbf{I})}[\log(1 - D_\psi(G_\theta(\mathbf{z})))]
\end{equation}

We implemented only the unconditional GAN variant due to convergence difficulties. Despite extensive efforts to condition the GAN on masked inputs (similar to our VAE approach), the conditional variant proved too unstable to converge on our complex multi-component dataset, necessitating focus on the unconditional formulation only.

\subsubsection{Variational Autoencoder (VAE)}

The VAE \cite{kingma2013auto} formulates generative modeling as probabilistic inference. The joint distribution factorizes into a Gaussian prior $p(\mathbf{z}) = \mathcal{N}(\mathbf{0}, \mathbf{I})$ and conditional likelihood $p_\theta(\bar{\mathbf{x}}|\mathbf{z})$. Training maximizes the Evidence Lower Bound (ELBO):
\begin{equation}
\mathcal{L}_{\text{VAE}} = \mathbb{E}_{q_\phi(\mathbf{z}|\bar{\mathbf{x}})}[\log p_\theta(\bar{\mathbf{x}}|\mathbf{z})] - \beta \text{KL}(q_\phi(\mathbf{z}|\bar{\mathbf{x}}) \parallel p(\mathbf{z}))
\end{equation}

To enable conditional generation, we implement a progressive masking strategy during training (see Appendix~\ref{app:vae-masking} for complete details).

\subsubsection{Multimodal Variational Autoencoder Plus (MMVAE+)}

The MMVAE+ \cite{palumbo2023mmvae+} extends the VAE framework by introducing structured latent codes with both shared and component-specific representations. It decomposes the latent space into a global latent variable $\mathbf{z}$ and component-specific latent variables $\mathbf{w}_i$ for each component $\mathbf{x}_i$. 

The model aggregates the shared latent code as a mixture-of-experts:
\begin{equation}
q_\phi(\mathbf{z}|\mathbf{X}) = \frac{1}{M} \sum_{i=1}^{M} q_{\phi_i}(\mathbf{z}|\mathbf{x}_i)
\end{equation}

And the approximate posterior factorizes as:
\begin{equation}
q_\phi(\mathbf{z},\mathbf{W}|\mathbf{X}) = q_\phi(\mathbf{z}|\mathbf{X}) \prod_{i=1}^{M} q_\phi(\mathbf{w}_i|\mathbf{x}_i)
\end{equation}

Training employs stratified sampling \cite{shi2019variational} to approximate the intractable ELBO:

{\small
\begin{equation}
\mathcal{L}_{\text{MMVAE}} = \frac{1}{M} \sum_{i=1}^{M} 
\mathbb{E}_{\substack{q_{\phi_i}(\mathbf{z}|\mathbf{x}_i) \\ q_{\phi}(\mathbf{W}|\mathbf{X})}} 
\left[ \log \frac{p_{\theta}(\mathbf{X}, \mathbf{z}, \mathbf{W})}
{q_{\phi}(\mathbf{z}|\mathbf{X})\, q_{\phi}(\mathbf{W}|\mathbf{X})} \right]
\end{equation}}

The MMVAE+ architecture inherently supports conditional generation since it learns to reconstruct all components given any individual component. During inference, we can simply provide one component as input and sample the remaining components from the learned conditional distribution.

\subsubsection{Denoising Diffusion Probabilistic Model (DDPM)}

The DDPM \cite{ho2020denoising} defines a forward diffusion process that progressively adds Gaussian noise to the data according to a predefined schedule. The forward process is defined by the conditional distribution:
\begin{equation}
q(\mathbf{x}_t|\mathbf{x}_{t-1}) = \mathcal{N}(\mathbf{x}_t; \sqrt{\alpha_t}\mathbf{x}_{t-1}, (1 - \alpha_t)\mathbf{I})
\end{equation}

The reverse denoising process is parameterized by a neural network $\epsilon_\theta(\mathbf{x}_t, t)$ that predicts the noise added at each step. The model is trained by minimizing:
\begin{equation}
\mathcal{L}_{\text{DDPM}} = \mathbb{E}_{t,\mathbf{x}_0,\epsilon} \left[ \| \epsilon - \epsilon_\theta(\mathbf{x}_t, t) \|^2 \right]
\end{equation}

For conditional generation, we implement diffusion-based inpainting. During the reverse sampling process, we maintain the observed components fixed while allowing the missing components to be generated. 

\begin{figure*}[t]
    \centering
    \includegraphics[width=.9\textwidth]{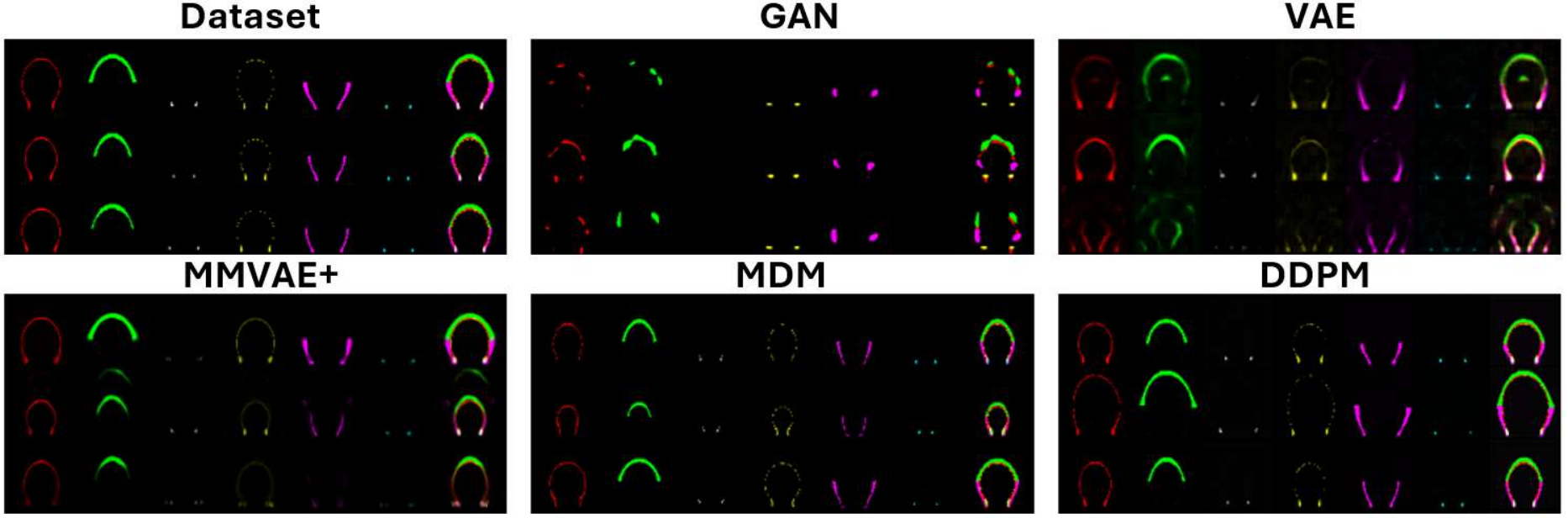}
    \caption{Qualitative results of unconditional generations for the 5 benchmarked DGMs. For each model, the first five columns represent the generated components, and the last column shows their superposition. The images are colored for easier visualization.}
    \label{fig:uncond_gen}
\end{figure*}

\begin{figure*}[t]
    \centering
    \includegraphics[width=.85\textwidth]{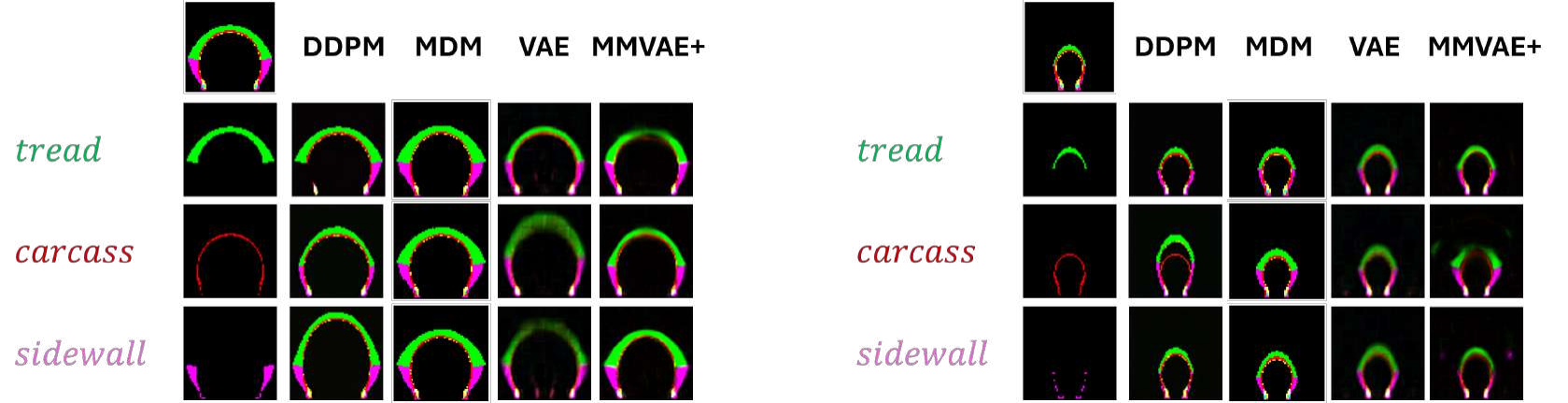}
    \caption{Qualitative results of component-constrained conditional generation. The figure shows two examples (top and bottom blocks). In each block, the leftmost column contains components from the test set, with the first image showing the assembled components forming a complete tire architecture. The remaining columns show the full tire architectures generated conditionally from a single component (from the first column). Each model (VAE, MMVAE+, DDPM, and MDM) generates complete architectures by either inpainting or conditional generation methods.}
    \label{fig:cond_gen}
\end{figure*}

\subsubsection{Multinomial Diffusion Model (MDM)}

The MDM \cite{hoogeboom2021argmax} extends the diffusion framework to
categorical data, it operates directly on the one-hot label tensor
$\mathbf c$ defined in Section \ref{sec:methods:notation}.

The forward diffusion process is governed by a categorical kernel that gradually transforms the data distribution toward a uniform categorical distribution:
\begin{equation}
q(\mathbf{c}_t \mid \mathbf{c}_{t-1})
    = \mathrm{Cat}\!\bigl(
        (1-\beta_t)\,\mathbf{c}_{t-1}
        + \beta_t/K \bigr),
\end{equation}
where $K=M{+}1$  and
$\mathbf{u}$ is the uniform probability vector.

The reverse denoising step uses a neural network that, given $(\mathbf c_t,t)$, 
predicts per-pixel logits for an estimate $\hat{\mathbf c}_0$.  
These logits are plugged into the closed-form posterior  
$q(\mathbf c_{t-1}\!\mid\mathbf c_t,\hat{\mathbf c}_0)$ to obtain  
$p_\theta(\mathbf c_{t-1}\!\mid\mathbf c_t)$, exactly as Eq.\,(14) in  
\cite{hoogeboom2021argmax} .


\paragraph{Categorical inpainting}
Analogous to the Gaussian “delta–mask’’ trick of \cite{sohl2015deep},
we multiply every reverse step by a \emph{binary mask}
that freezes the observed labels.
Let $\mathbf c_0\!\in\!\{0,1\}^{K \times H\times W}$ and
$\mathcal O$ be the set of observed pixels.
For each pixel we set the mask
$r_k = (\mathbf{c}_0^i)_k$, $k\in\{1,\dots,K\} $ if $i \in \mathcal O$
and $r_k = 1$ otherwise.

The resulting distribution remains categorical as per Proposition\ref{prop:masked_cat}, sampling therefore costs no more than in the unconditional case.

\begin{proposition}[Masked categorical reverse step]
\label{prop:masked_cat}
Let $p(\mathbf{c}_{t-1}|\mathbf{c}_t)=\Cat(\mathbf{c}_{t-1}|\boldsymbol{\theta}(\mathbf{c}_t, \mathbf{c}_{0}))$ (Equation (14) in \cite{hoogeboom2021argmax}) be the learned reverse-diffusion kernel at time $t$, where $\boldsymbol{\theta} = (\theta_0,\ldots,\theta_K)$ satisfies $\theta_k \geq 0$ and $\sum_{k=0}^{K}\theta_k = 1$.

Let $\mathbf{r} = (r_1,\ldots,r_K)$ be any non-negative weight vector with $\sum_{k} r_k \theta_k > 0$.

Define the masked kernel:
\begin{equation}
\tilde{p}(\mathbf{c}_{t-1}|\mathbf{c}_t) = \frac{\mathbf{r(\mathbf{c}_{t-1})} \cdot p(\mathbf{c}_{t-1}|\mathbf{c}_t)}{\sum_{j=1}^{K} r_j \cdot \theta_j}
\end{equation}

Then $\tilde{p}(\mathbf{c}_{t-1}|\mathbf{c}_t)$ is again a categorical distribution:
\begin{equation}
\tilde{p}(\mathbf{c}_{t-1}|\mathbf{c}_t) = \Cat(\mathbf{c}_{t-1}|\tilde{\boldsymbol{\theta}}), \quad \tilde{\theta}_k = \frac{r_k \cdot \theta_k}{\sum_{j=0}^{K} r_j \cdot \theta_j}
\end{equation}
\end{proposition}

Where $\mathbf{r}(\mathbf{c}_{t-1})$ denotes the scalar product $\langle \mathbf{r}, \mathbf{c}_{t-1} \rangle $.


\subsubsection{Dimension-Constrained Generation} 
For dimension-constrained generation, we adopt classifier-free guidance \cite{ho2022classifier} with both DDPM and MDM models, enabling them to generate tire architectures that satisfy specific dimensional constraints without training an auxiliary classifier.

\section{Experiments}
\label{sec:experiments}
This section evaluates the five generative models described previously across three industrially relevant generation scenarios, using metrics specifically designed to reflect key tire manufacturing requirements. We first detail our experimental setup and evaluation metrics, then present results for each scenario in sequence: unconstrained generation, component-conditioned generation (emphasizing conditions most relevant to industrial design), and dimension-constrained generation. Given the clear superiority of diffusion-based models observed in unconstrained generation, we limit our evaluation of dimension-constrained generation to DDPM and MDM architectures, allowing a more focused assessment of their capabilities.

To ensure robust metric evaluation, we generated a sufficient number of samples in each scenario. For unconstrained generation, we generated 10,000 samples for comprehensive metric computation, including FID. For component-conditioned generation, we produced 45 samples per conditioning scenario, collectively totaling over 10,000 samples. In dimension-constrained generation, the same sampling procedure was applied for in-distribution scenarios, while for out-of-distribution cases, we generated 1,000 samples and omitted FID scores, as comparison against in-distribution reference data would not yield meaningful perceptual quality insights.

All proposed metrics and evaluation procedures are dataset-agnostic and directly transferable to other industrial design problems involving similar multi-component structures.

\subsection{Experimental Setup}

\paragraph{Model implementation}
All models are implemented with comparable parameter counts (Table \ref{tab:model_complexity}) and trained for 100 epochs. Due to training instabilities, GANs were trained for up to 1000 epochs with checkpointing to retain the best-performing model. Architectural and hyperparameter details are provided in Appendix \ref{app:architectures}.

\paragraph{Evaluation metrics}
We employ complementary evaluation criteria designed to assess different aspects of generative quality. These metrics are selected for their relevance to industrial applications while remaining generalizable to other multi-component systems.

\textbf{Component overlap (IoU).}
For each of the $\binom{M}{2}=15$ unordered component pairs, empirical distributions of Intersection-over-Union (IoU) \cite{rezatofighi2019generalized} are computed over the dataset and over the generated set. A separate $W_1$ distance is calculated for each pair. Averaging these 15 distances yields a single score (which we refer to as the IoU metric for brevity) in the unconditional setting, and one score per conditioning component in the component-conditioned setting (Table~\ref{tab:iou}). For clarity, Figure~\ref{fig:all_metrics} shows the mean of these component-wise scores. The Multinomial Diffusion Model (MDM) is excluded from this evaluation since its categorical output forbids pixel overlap by construction.

\textbf{Spatial positioning - Center of Mass (CoM).} 
For each sample, the two-dimensional center of mass (CoM) of every component is computed, the twelve coordinates are concatenated into a single 12-dimensional vector, and the $W_1$ between the joint real-versus-generated distributions is taken as the CoM metric. For component-conditioned generation, the metric is evaluated separately for each conditioning component (Table \ref{tab:iou}); the figure in the main text displays the mean across conditioning components to streamline comparison.

\textbf{Region-Connectivity Error (RCE).} In real tire architectures, all components assemble into a single contiguous region. Connectivity fidelity is quantified by counting connected regions in the assembled mask using connected component analysis (8-neighbor connectivity) and computing the 1-Wasserstein distance to the ideal Dirac distribution $\delta_1$ that places all probability mass at exactly one connected component, where lower values indicate lower fragmentation.

\textbf{Fréchet Inception Distance (FID).} 
Overall visual fidelity is assessed using the Fréchet Inception Distance (FID) between real and generated images of the assembled tire architectures. This metric evaluates the similarity between feature distributions extracted by a pre-trained Inception network.

\textbf{Dimension Error (DimErr).} 
For the dimension-constrained generation task, adherence to specified dimensional requirements is quantified using the absolute deviation between the target dimensions and the actual dimensions of the generated tire architecture. Formally, for a generated sample $\mathbf{x}$ conditioned on target dimensions $\mathbf{d}_{\text{target}} = (w_{\text{target}}, h_{\text{target}})$, we measure its actual dimensions $\mathbf{d}_{\text{actual}} = (w_{\text{actual}}, h_{\text{actual}})$ and compute:
\begin{align}
\text{DimErr}(\mathbf{x}) 
  &= \bigl|w_{\text{target}} - w_{\text{actual}}\bigr|
     + \bigl|h_{\text{target}} - h_{\text{actual}}\bigr| \label{eq:dimerr}
\end{align}

The final DimErr metric is computed as the average over all generated samples in the test set. 

\noindent For all metrics, lower values indicate better performance.

\subsection{Unconstrained Generation}
\begin{figure}[!t]
    \centering
    \includegraphics[width=0.5\textwidth]{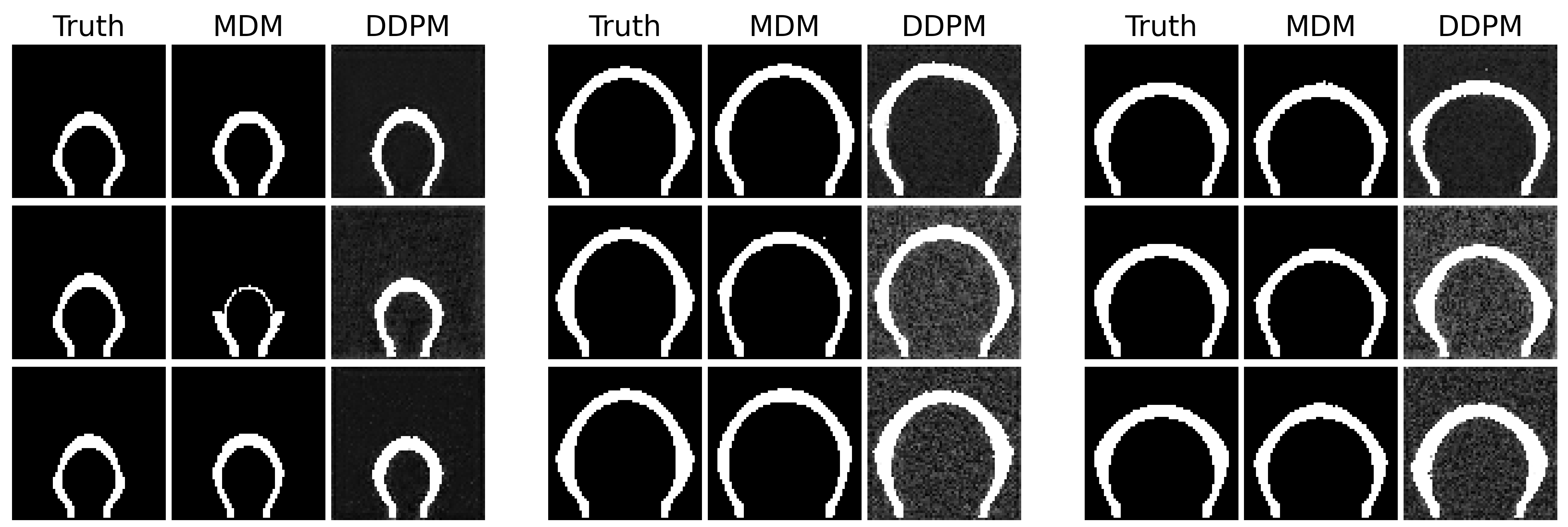}
    \caption{In-distribution comparison of dimension-constrained generation. Each block shows three columns: the test image (left), followed by three samples from MDM (middle), and three samples from DDPM (right). From left to right across the 3 blocks, the target dimensions $(w_{\text{target}}, h_{\text{target}})$ are $(30, 35)$, $(58, 54)$, and $(56, 47)$.}
    \label{fig:sample_comparison_ID}
\end{figure}

\subsubsection{Qualitative assessment}
Figure~\ref{fig:uncond_gen} shows representative samples from each model. GAN produces sharp but severely fragmented images with limited variability, exhibiting classic mode collapse symptoms. VAE generates blurry images with duplicate components at different scales, attributable to the prior-hole problem in data-limited settings \cite{bauer2019resampled}. MMVAE+ improves image sharpness over VAE but retains duplication artifacts. In contrast, diffusion-based models (DDPM and MDM) generate clean, visually plausible tire architectures with coherent inter-component relationships and no duplication artifacts.
Due to GAN's poor performance and convergence issues despite extensive hyperparameter tuning, we exclude this approach from subsequent analyses.


\begin{figure*}[t]  
  \centering
  \setlength{\belowcaptionskip}{-15pt}  
  
  \begin{minipage}{0.95\textwidth}
    \begin{minipage}{0.04\textwidth}
      \centering
      \raisebox{1.05cm}{\rotatebox{90}{IoU}}
    \end{minipage}%
    \begin{minipage}{0.22\textwidth}
      \includegraphics[width=\textwidth]{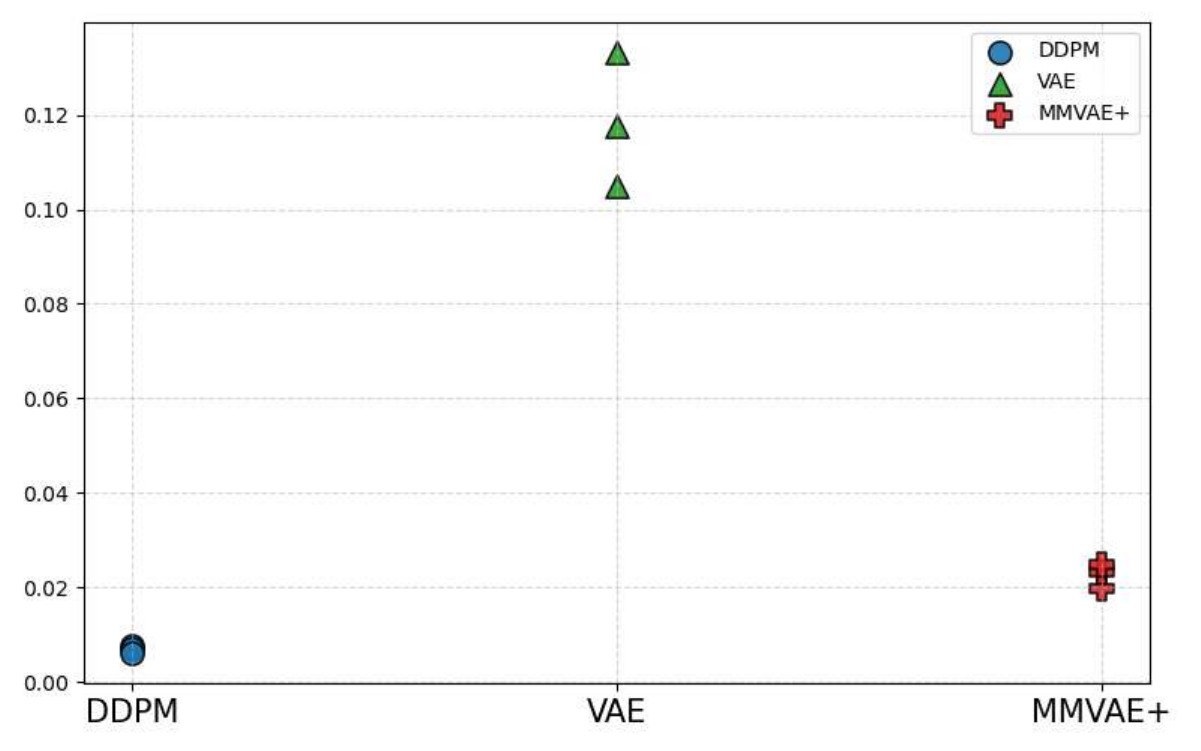}
      \centering
      \scriptsize{Unconstrained Generation}
    \end{minipage}%
    \hfill
    \begin{minipage}{0.22\textwidth}
      \includegraphics[width=\textwidth]{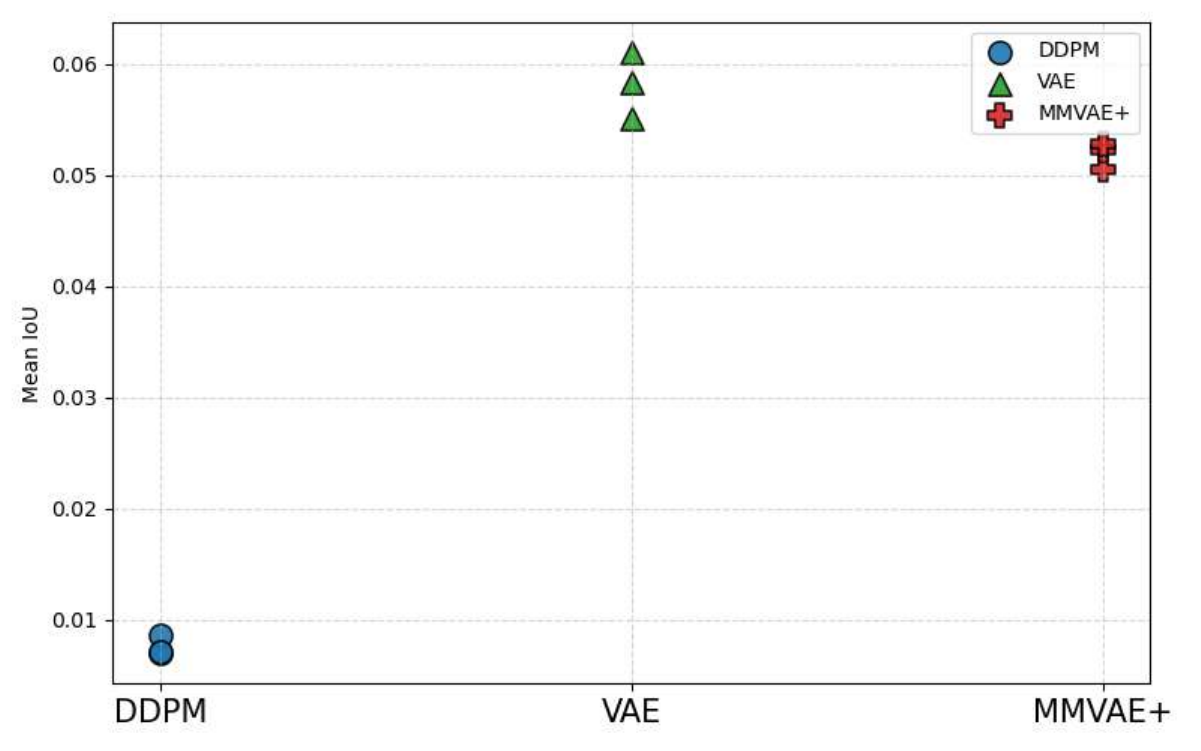}
      \centering
      \scriptsize{Component-Conditioned}
    \end{minipage}
    \hfill
    \begin{minipage}{0.04\textwidth}
      \centering
      \raisebox{1.05cm}{\rotatebox{90}{CoM}}
    \end{minipage}%
    \begin{minipage}{0.22\textwidth}
      \includegraphics[width=\textwidth]{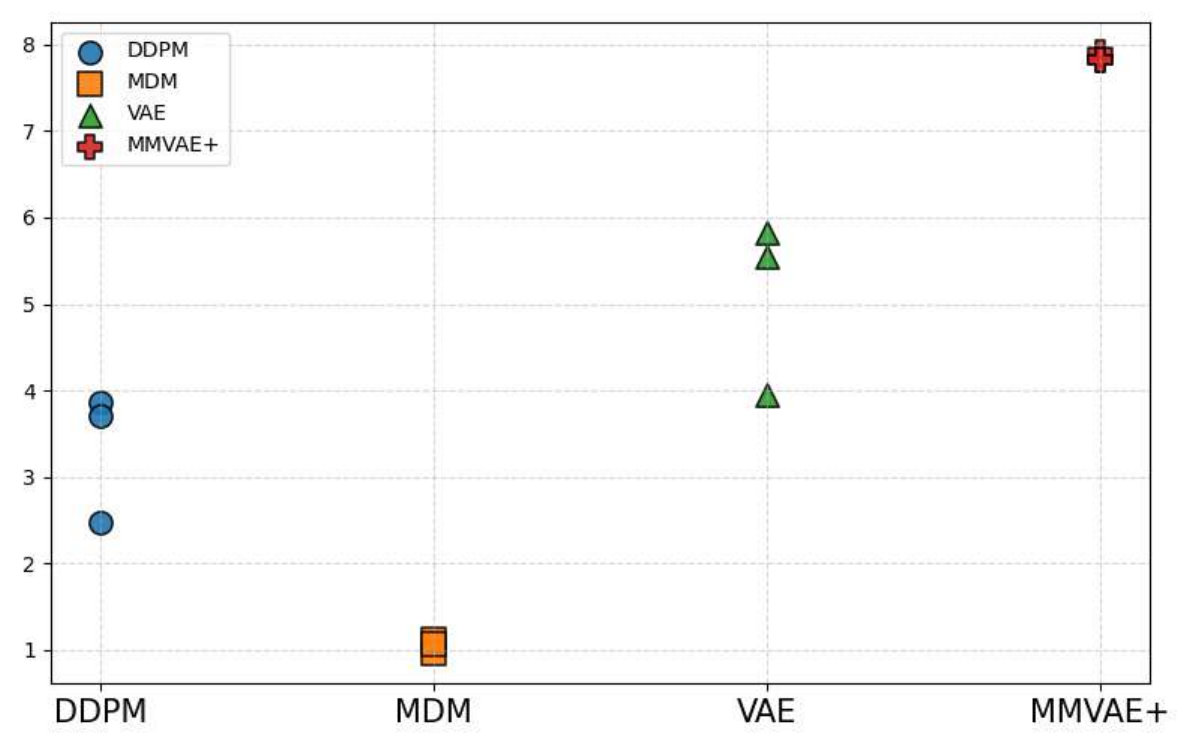}
      \centering
      \scriptsize{Unconstrained Generation}
    \end{minipage}%
    \hfill
    \begin{minipage}{0.22\textwidth}
      \includegraphics[width=\textwidth]{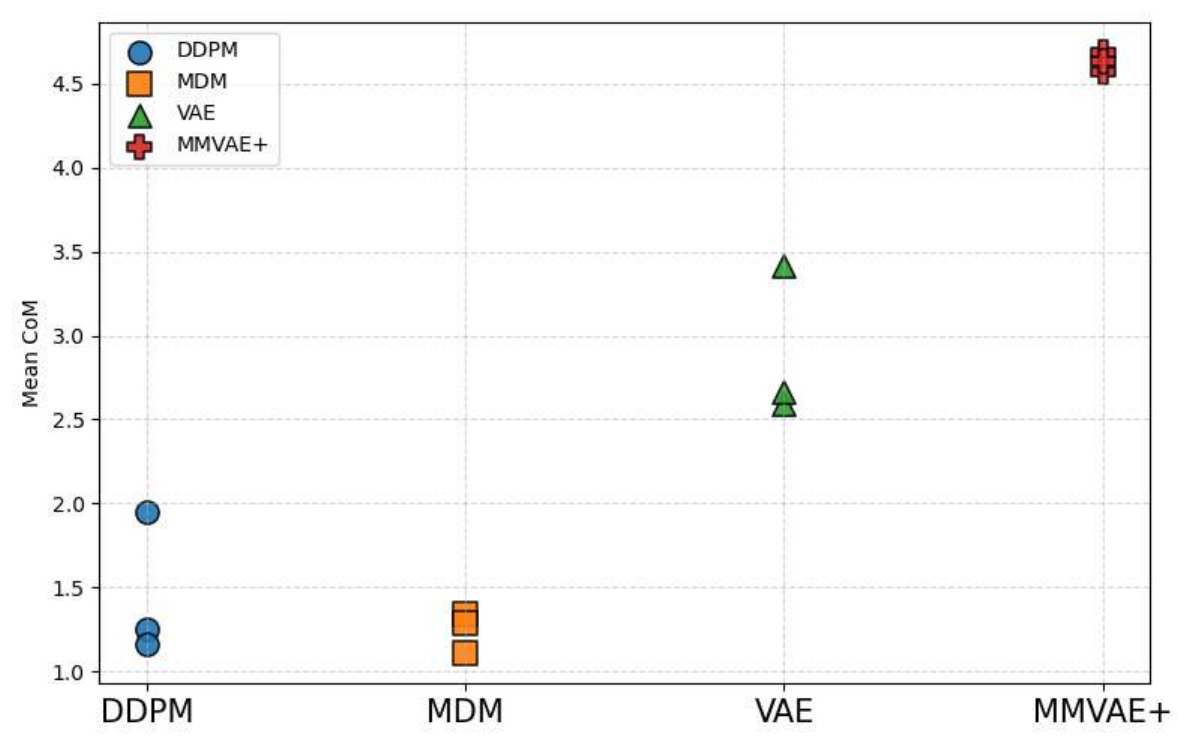}
      \centering
      \scriptsize{Component-Conditioned}
    \end{minipage}
  \end{minipage}

  \vspace{0.7em} 

  \begin{minipage}{0.95\textwidth}
    \begin{minipage}{0.04\textwidth}
      \centering
      \raisebox{1.05cm}{\rotatebox{90}{RCE}}
    \end{minipage}%
    \begin{minipage}{0.22\textwidth}
      \includegraphics[width=\textwidth]{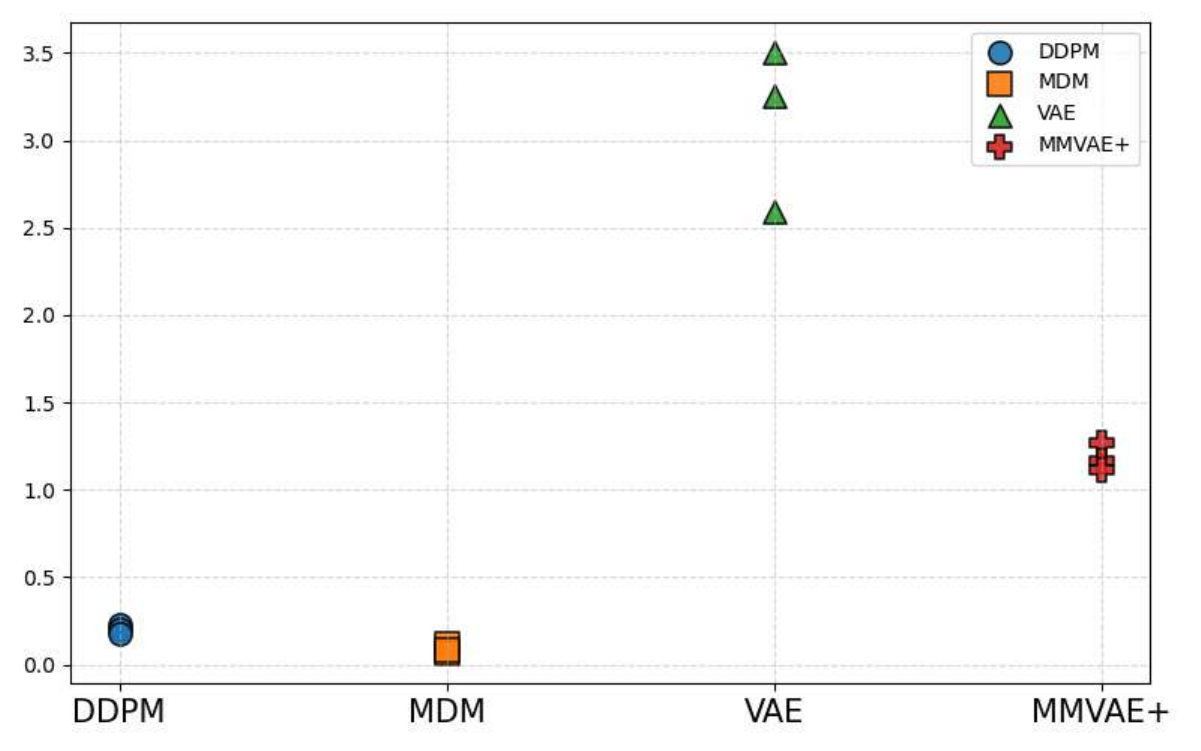}
      \centering
      \scriptsize{Unconstrained Generation}
    \end{minipage}%
    \hfill
    \begin{minipage}{0.22\textwidth}
      \includegraphics[width=\textwidth]{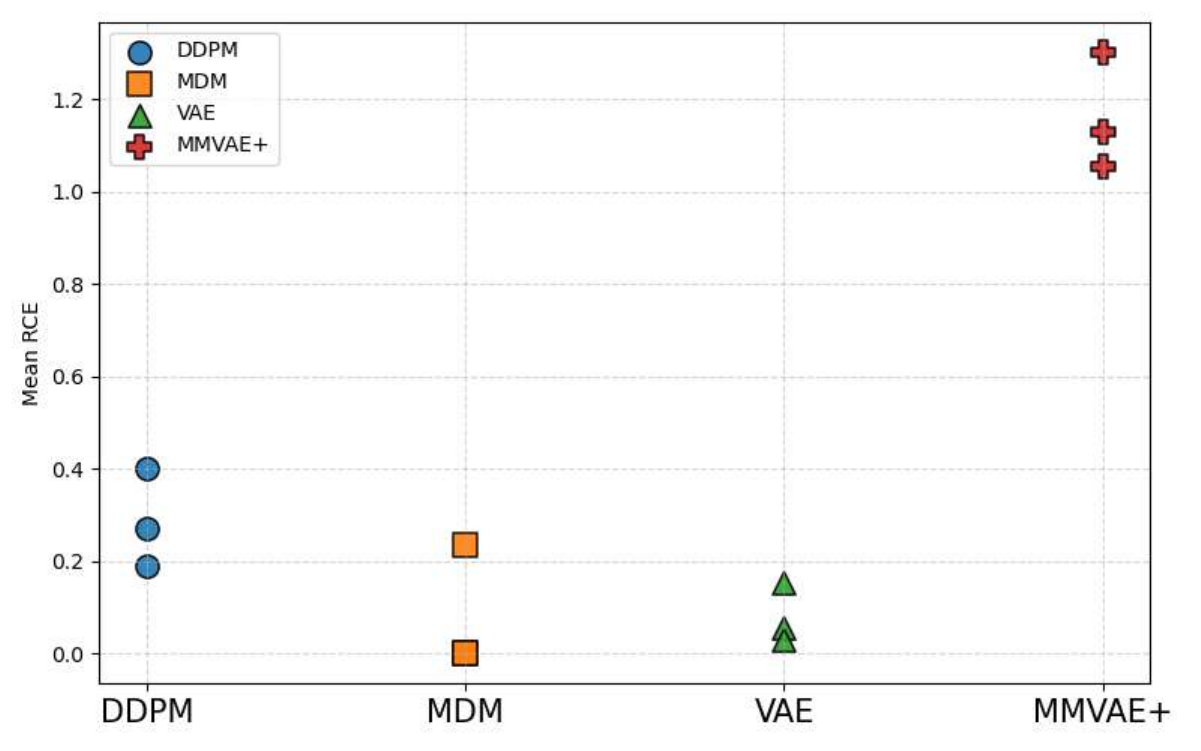}
      \centering
      \scriptsize{Component-Conditioned}
    \end{minipage}
    \hfill
    \begin{minipage}{0.04\textwidth}
      \centering
      \raisebox{1.05cm}{\rotatebox{90}{FID}}
    \end{minipage}%
    \begin{minipage}{0.22\textwidth}
      \includegraphics[width=\textwidth]{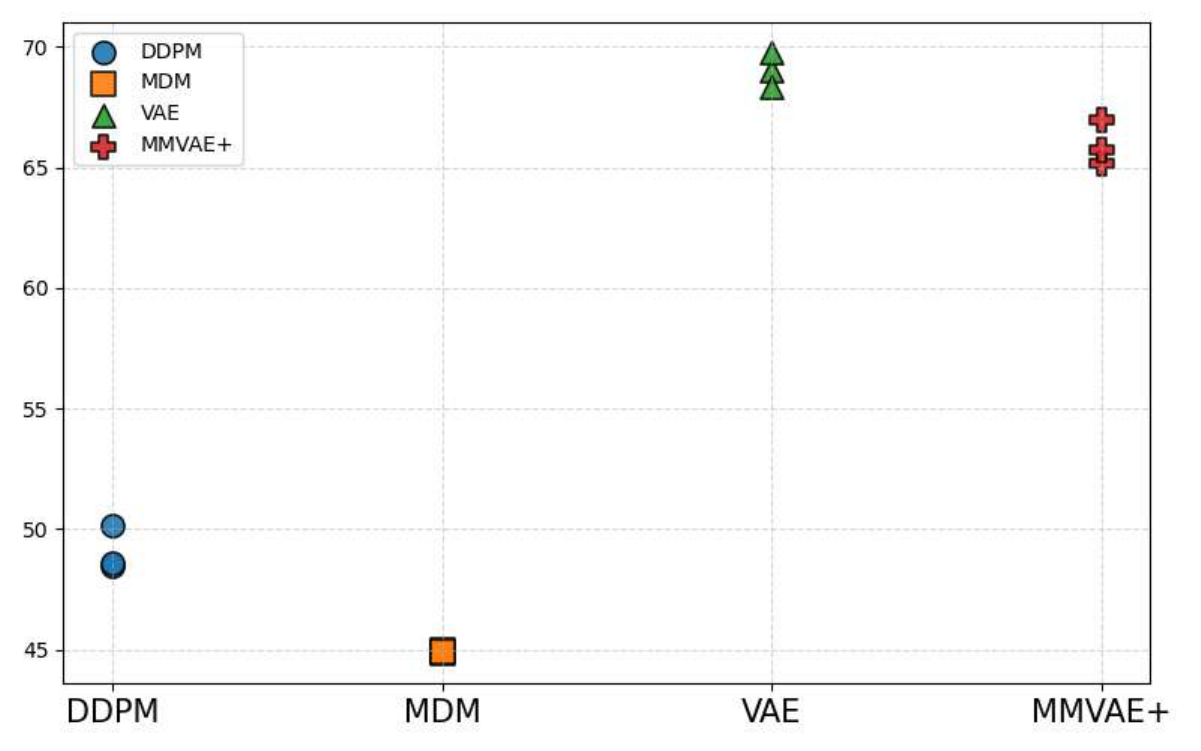}
      \centering
      \scriptsize{Unconstrained Generation}
    \end{minipage}%
    \hfill
    \begin{minipage}{0.22\textwidth}
      \includegraphics[width=\textwidth]{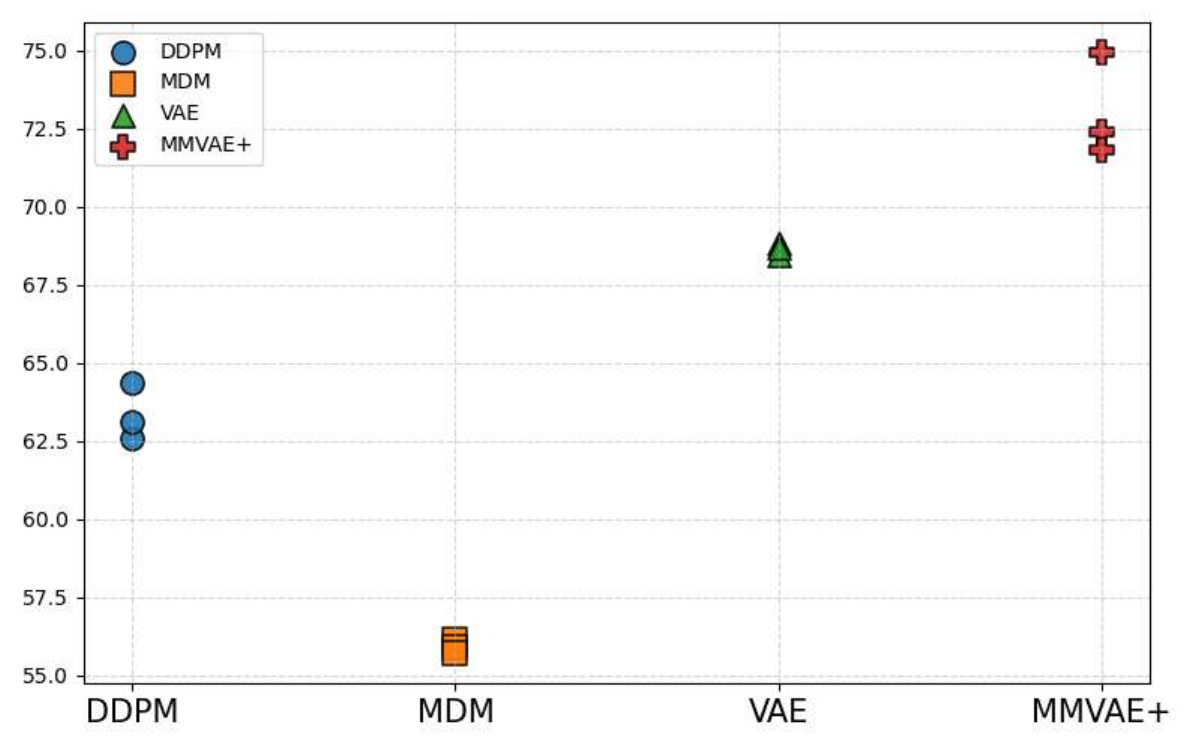}
      \centering
      \scriptsize{Component-Conditioned}
    \end{minipage}
  \end{minipage}
  \caption{Performance comparison of generative models across four metrics (IoU, CoM, RCE, FID) for unconstrained generation and component-conditioned generation. Lower values indicate better performance. Points represent individual training runs with different random seeds. Diffusion-based models (MDM, DDPM) consistently outperform in unconstrained tasks, while standard VAE shows competitive connectivity (RCE) in component-conditioned scenarios.}
  \label{fig:all_metrics}
  \end{figure*}
  
\subsubsection{Quantitative results}
Our comprehensive evaluation across five complementary metrics (Figure \ref{fig:all_metrics}) reveals a consistent performance hierarchy among the benchmarked models. MDM demonstrates superior performance in all applicable metrics (CoM, RCE, and FID), with DDPM following as a strong second-best performer and leading in IoU where MDM is not evaluated due to its categorical nature. MMVAE+ generally outperforms VAE in most metrics except spatial coherence (CoM), where VAE shows slightly better component positioning.

These objective findings align with our qualitative observations, establishing diffusion-based models as the most promising approaches for industrial tire architecture generation. The substantial performance gap between diffusion models and their VAE/GAN counterparts can be attributed to the efficacy of iterative denoising processes. This stepwise refinement allows diffusion models to progressively resolve complex spatial relationships and maintain structural integrity across multi-component industrial designs, whereas single-pass generative approaches must capture these intricate dependencies in a single transformation.

\subsection{Component-Conditioned Generation}

\subsubsection{Qualitative assessment}
Figure \ref{fig:cond_gen} shows that component conditioning significantly enhances the fidelity and coherence of generations for VAE and MMVAE+. Notably, the standard VAE produces samples with superior structural integrity compared to MMVAE+, characterized by sharper boundaries, fewer artifacts, and better component connectivity. This quality even surpasses DDPM in specific aspects, as the diffusion model occasionally generates topologically inconsistent outputs with disconnected components.


\subsubsection{Quantitative assessment}


The model ranking between diffusion-based and VAE-based approaches across metrics remains largely consistent with the unconditional case, with diffusion-based models generally maintaining superior performance. However, an important exception emerges in the RCE metric, where VAE demonstrates competitive results approaching MDM's performance and surpassing other models. 
The metrics show that VAE outperforms MMVAE+ on most evaluated aspects. This behavior might stem from architectural and training differences: the VAE ingests a single six-channel tensor and, through the progressive-mask curriculum, learns strong cross-component priors that reconstruct missing parts from partial input. MMVAE+, in contrast, encodes each component separately and fuses posteriors via a mixture-of-experts, limiting early feature sharing when only one component is supplied. A more detailed analysis is provided in Appendix \ref{app:vae_vs_mmvae}.


\begin{table}[t]
    \centering

    \begin{tabular}{lcc}
        \toprule
        Model & Parameters (M) & GFLOPs \\
        \midrule
        VAE & 11.47 & 3.43 \\
        GAN (Generator) & 13.49 & 4.21\\
        GAN (Discriminator) & 7.43 & 2.95 \\
        MMVAE+ & 9.67 & 18.77 \\
        DDPM & 9.91 & 11.96 \\
        MDM & 9.64 & 9.87 \\
        \bottomrule
    \end{tabular}
\caption{Model Parameters and Computational Complexity}
\label{tab:model_complexity}
\end{table}

\subsection{Dimension-Constrained Generation}

\subsubsection{Qualitative assessment}
Figure \ref{fig:sample_comparison_ID} presents dimension-constrained samples generated by both MDM and DDPM models when conditioned on in-distribution (ID) dimensional parameters. Visual inspection reveals that both models generally adhere to the specified dimensional constraints, though DDPM-generated samples exhibit higher noise levels. 

\begin{figure*}[t]
    \centering
    \begin{minipage}{0.29\textwidth}  
        \centering
        \includegraphics[width=\textwidth]{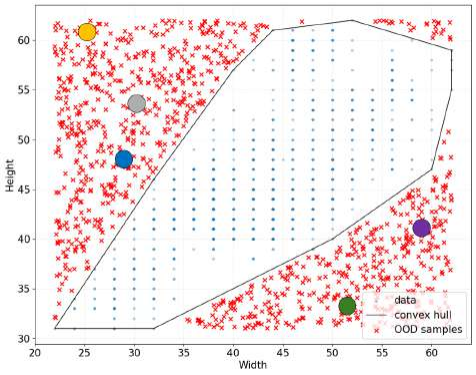}
    \end{minipage}%
    \hfill
    \begin{minipage}{0.69\textwidth}  
        \centering
        \includegraphics[width=\textwidth]{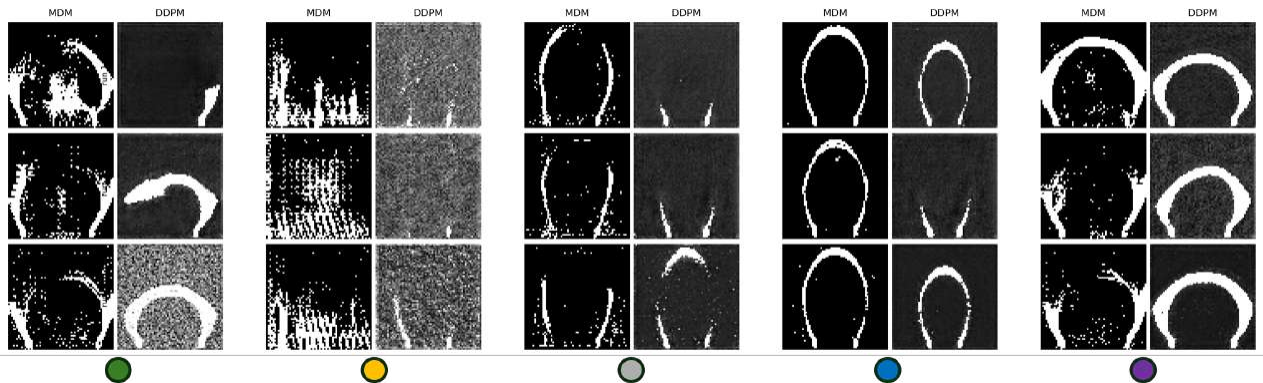}
    \end{minipage}
    \caption{OOD sample comparison of dimension-constrained generation. The left plot shows the distribution hull from Figure \ref{fig:id_ood_hull}, with colored circles indicating the specific dimensional constraints used for each of the eight test cases. The right portion shows samples from MDM (left column) and DDPM (right column) for each dimensional constraint. Each column contains three different samples generated with the same width and height constraints, demonstrating how both models extrapolate to generate tire architectures with dimensions beyond the training distribution.}
    \label{fig:sample_comparison_OOD}
\end{figure*}

For out-of-distribution (OOD) scenarios (Figure \ref{fig:sample_comparison_OOD}), we observe a negative correlation between generation quality and distance from the training distribution's convex hull. Samples conditioned on dimensions farther from the observed distribution boundary display more artifacts and structural inconsistencies. Notably, DDPM demonstrates superior extrapolation capabilities in the OOD domain, producing more plausible architectures under novel dimensional constraints compared to MDM, despite its higher noise characteristics in the ID domain. This suggests a potential trade-off between precision within the training distribution and generalization capacity beyond it.

\subsubsection{Quantitative assessment}
\begin{figure}[!t]
  \centering

  \begin{subfigure}[t]{\columnwidth}
    \begin{minipage}[c]{0.78\columnwidth}
      \includegraphics[width=1.\linewidth]{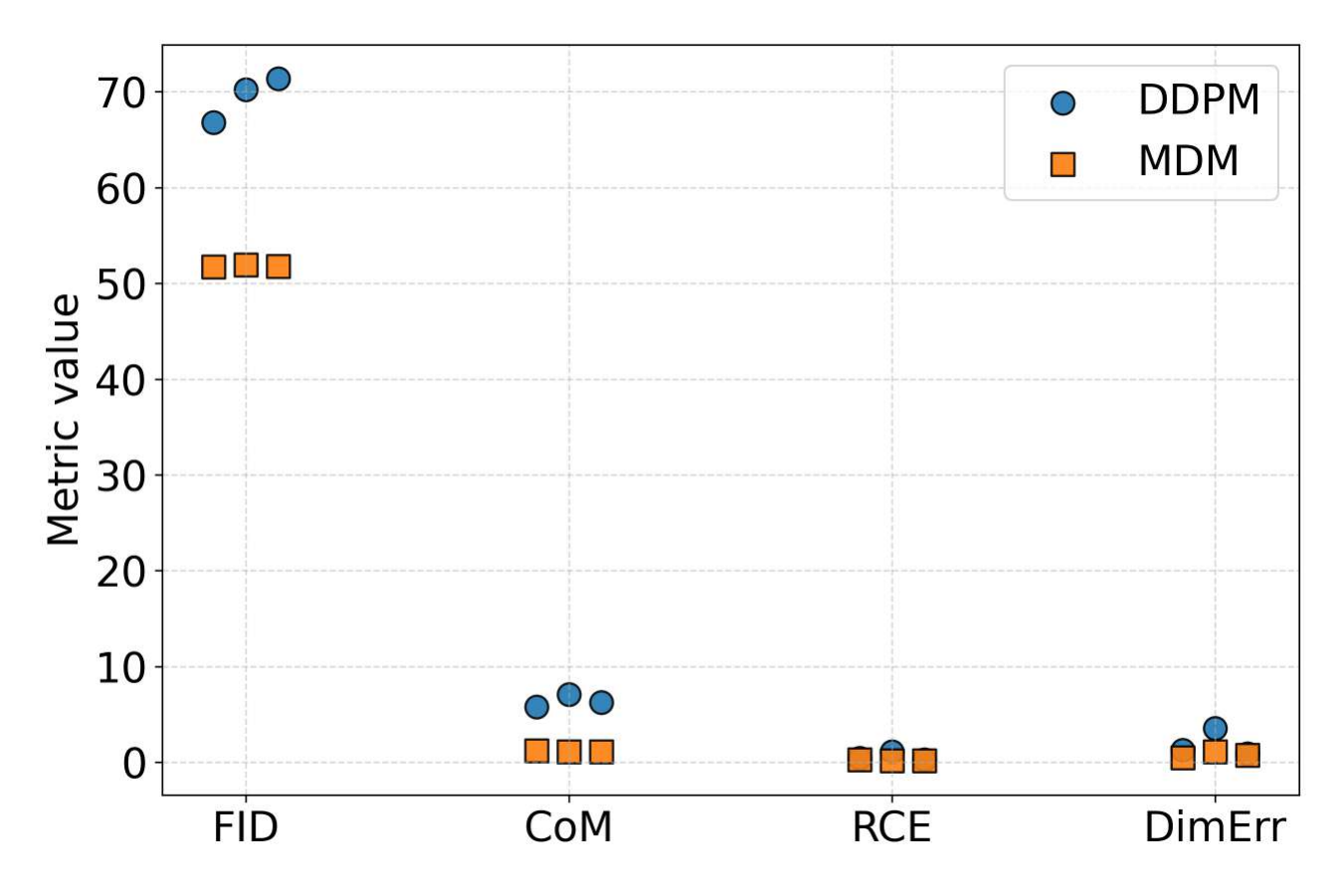}
    \end{minipage}%
    \hfill
    \begin{minipage}[c]{0.2\columnwidth}
      \caption{ID}\label{subfig:id}
    \end{minipage}
  \end{subfigure}

  \vspace{6pt}

  \begin{subfigure}[t]{\columnwidth}
    \begin{minipage}[c]{0.78\columnwidth}
      \includegraphics[width=1.\linewidth]{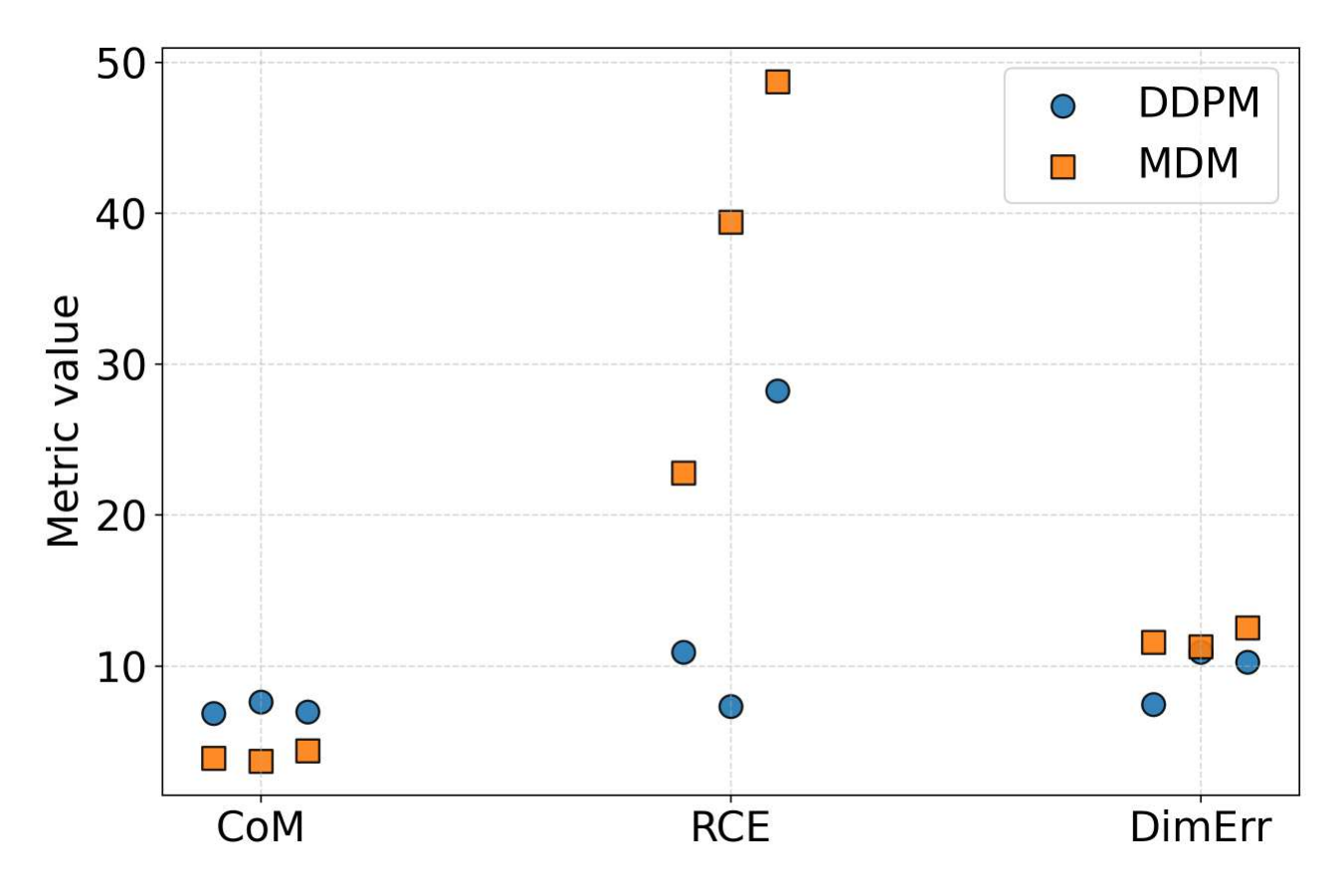}
    \end{minipage}%
    \hfill
    \begin{minipage}[c]{0.20\columnwidth}
      \caption{OOD}\label{subfig:ood}
    \end{minipage}
  \end{subfigure}

  \caption{Quantitative performance of diffusion models on
           dimension-constrained generation.  Metrics:
           FID, CoM, RCE, and dimension error (ID);
           CoM, RCE, and dimension error (OOD).}
  \label{fig:dim_metrics}
\end{figure}

Our quantitative evaluation of dimension-constrained generation compares model performance on both in-distribution and out-of-distribution test cases, as shown in Figure \ref{fig:dim_metrics}. For the ID scenario, we evaluate models the four metrics: FID, CoM, RCE, and DimErr. For the OOD scenario, we exclude FID as a metric since comparing out-of-distribution samples against in-distribution data no longer provides meaningful information about perceptual quality.

The quantitative results corroborate our qualitative observations. In the ID domain, MDM outperforms DDPM across most metrics, particularly in FID, which aligns with the noisier images observed in DDPM-generated samples.

For OOD scenarios, DDPM consistently outperforms MDM in RCE, as evidenced in Figure \ref{fig:sample_comparison_OOD}, where DDPM better preserves component connectivity, even at the cost of generating asymmetrical tire architectures. This asymmetry negatively impacts the CoM metric for DDPM. This performance pattern highlights a critical trade-off: MDM performs better  at reproducing tire architectures similar to historical designs with higher visual fidelity, while DDPM demonstrates superior extrapolative capabilities for novel dimensional specifications. This finding has important implications for industrial deployment, suggesting that model selection should be driven by the specific objectives: either prioritizing fidelity to existing designs or enabling exploration of novel architectural configurations beyond the training distribution.

\section{Conclusion}

This paper presents a comprehensive benchmark of deep generative models for industrial tire architecture design, evaluated qualitatively and quantitatively across unconditional generation, component-conditioned generation, and dimension-constrained generation scenarios.

Diffusion-based models consistently outperform other approaches. The choice between MDM and DDPM depends on specific use cases: MDM generates higher-fidelity samples within the training distribution, while DDPM offers superior extrapolation to novel dimensional specifications. Interestingly, simpler architectures like the VAE can surpass specialized models MMVAE+ in conditional generation tasks.

These insights represent the core scientific contribution of this work, independent of the dataset specifics, as our framework, metrics, and generative strategies are transferable to other multi-component design problems.

Future work will explore multimodal diffusion approaches to better model component interactions and develop perceptual quality metrics suited to out-of-distribution generations, addressing limitations of current metrics such as FID.

\newpage
\bibliographystyle{IEEEtran}
\bibliography{references}

\newpage
\appendix
\subsection{Detailed Results Tables}


Tables~\ref{tab:fid}--\ref{tab:metrics_id_ood} report all quantitative
results\footnote{%
Table~\ref{tab:metrics_id_ood} spans both columns and may float to the
next page.}. 
Every entry is the \emph{mean $\pm$ standard
deviation} over three independent training runs with distinct random seeds.

\begin{table}[H]
\centering
\begin{tabular}{lcc}
\toprule
\textbf{Model} & \textbf{Uncond.} & \textbf{Cond.} \\
\midrule
DDPM     & 49.1 $\pm$ 1.0 & 63.3 $\pm$ 0.9 \\
MDM      & 44.9 $\pm$ 0.1 & 55.9 $\pm$ 0.2 \\
VAE      & 69.0 $\pm$ 0.7 & 68.6 $\pm$ 0.2 \\
MMVAE+   & 65.9 $\pm$ 0.9 & 73.0 $\pm$ 1.7 \\
\bottomrule
\end{tabular}
\caption{FID metric for unconditional and component-conditioned generations (lower is better).}
\label{tab:fid}
\end{table}

\begin{table}[H]
\centering
\hspace*{-0.8cm} 
\begin{tabular}{lcccc}
\toprule
            & \textbf{Uncond.} & \multicolumn{3}{c}{\textbf{Conditional}} \\
            \cmidrule(lr){2-2}\cmidrule(lr){3-5}
\textbf{Model} &  & carcass & tread & sidewall \\
\midrule
DDPM     & 0.007 $\pm$ 0.001 & 0.004 $\pm$ 0.001 & 0.010 $\pm$ 0.001 & 0.008 $\pm$ 0.001 \\
VAE      & 0.119 $\pm$ 0.014 & 0.055 $\pm$ 0.002 & 0.058 $\pm$ 0.004 & 0.062 $\pm$ 0.004 \\
MMVAE+   & 0.023 $\pm$ 0.003 & 0.049 $\pm$ 0.001 & 0.047 $\pm$ 0.001 & 0.059 $\pm$ 0.002 \\
\bottomrule
\end{tabular}
\caption{IoU scores (lower is better).
Unconditional generation is summarised by a single IoU value, whereas the component‑conditioned setting reports one IoU score per conditioning component.}
\label{tab:iou}
\end{table}

\begin{table}[H]
\centering
\hspace*{-0.9cm}
\begin{tabular}{lcccc}
\toprule
            & \textbf{Uncond.} & \multicolumn{3}{c}{\textbf{Conditional}} \\
            \cmidrule(lr){2-2}\cmidrule(lr){3-5}
\textbf{Model} &  & carcass & tread & sidewall \\
\midrule
DDPM     & 3.34 $\pm$ 0.76 & 1.04 $\pm$ 0.61 & 0.75 $\pm$ 0.32 & 2.59 $\pm$ 1.53 \\
MDM      & 1.04 $\pm$ 0.08 & 2.32 $\pm$ 0.16 & 0.68 $\pm$ 0.20 & 0.74 $\pm$ 0.05 \\
VAE      & 5.10 $\pm$ 1.01 & 1.96 $\pm$ 1.50 & 2.27 $\pm$ 0.26 & 4.44 $\pm$ 2.54 \\
MMVAE+   & 7.85 $\pm$ 0.05 & 4.25 $\pm$ 0.04 & 1.82 $\pm$ 0.13 & 7.82 $\pm$ 0.10 \\
\bottomrule
\end{tabular}
\caption{CoM scores (lower is better) for unconditional and component‑conditioned generation.
For the component‑conditioned setting we report the metric separately for each component; Figure \ref{fig:all_metrics} presents their average for clarity.}
\label{tab:com}
\end{table}

\begin{table}[H]
\centering
\hspace*{-0.9cm}
\begin{tabular}{lcccc}
\toprule
            & \textbf{Uncond.} & \multicolumn{3}{c}{\textbf{Conditional}} \\
            \cmidrule(lr){2-2}\cmidrule(lr){3-5}
\textbf{Model} & overall & carcass & tread & sidewall \\
\midrule
DDPM     & 0.201 $\pm$ 0.024 & 0.426 $\pm$ 0.185 & 0.118 $\pm$ 0.091 & 0.319 $\pm$ 0.122 \\
MDM      & 0.087 $\pm$ 0.026 & 0.150 $\pm$ 0.260 & 0.047 $\pm$ 0.082 & 0.037 $\pm$ 0.064 \\
VAE      & 3.118 $\pm$ 0.470 & 0.012 $\pm$ 0.009 & 0.029 $\pm$ 0.018 & 0.197 $\pm$ 0.193 \\
MMVAE+   & 1.185 $\pm$ 0.076 & 0.949 $\pm$ 0.068 & 1.021 $\pm$ 0.187 & 1.516 $\pm$ 0.210 \\
\bottomrule
\end{tabular}
\caption{RCE scores (lower is better) for unconditional and component‑conditioned generation.
For the component‑conditioned setting we report the metric separately for each component; Figure \ref{fig:all_metrics} presents their average for clarity.}
\label{tab:rce}
\end{table}

\begin{table*}[h]
\centering
\begin{tabular}{lccccccc}
\toprule
& \multicolumn{4}{c}{\textbf{ID split}} & \multicolumn{3}{c}{\textbf{OOD split}}\\
\cmidrule(lr){2-5}\cmidrule(lr){6-8}
\textbf{Model} & FID $\downarrow$ & CoM $\downarrow$ & RCE $\downarrow$ & DimErr $\downarrow$ & CoM $\downarrow$ & RCE $\downarrow$ & DimErr $\downarrow$\\
\midrule
DDPM & 69.450 $\pm$ 2.363 & 6.396 $\pm$ 0.654 & 0.621 $\pm$ 0.455 & 1.905 $\pm$ 1.443 & 7.126 $\pm$ 0.417 & 15.485 $\pm$ 11.175 & 9.528 $\pm$ 1.843 \\
MDM & 51.798 $\pm$ 0.096 & 1.163 $\pm$ 0.058 & 0.204 $\pm$ 0.052 & 0.769 $\pm$ 0.301 & 3.986 $\pm$ 0.347 & 36.978 $\pm$ 13.138 & 11.795 $\pm$ 0.658 \\
\bottomrule
\end{tabular}
\caption{Metrics (\emph{mean $\pm$ std} over three runs) for the \textbf{dimension-constrained generations} illustrated in Fig.~\ref{fig:dim_metrics}.  
FID is reported only on the in-distribution (ID) split; lower values indicate better performance for every metric.}
\label{tab:metrics_id_ood}
\end{table*}

\subsection{VAE Progressive Masking Implementation}
\label{app:vae-masking}

To enable conditional generation with the VAE, we implement a progressive masking curriculum during training. Let $\bar{\mathbf{x}} \in \mathbb{R}^{C \times H \times W}$ denote a multi-component sample with $C = 6$ channels. At each epoch $e$, we define the masking probability:
\begin{equation}
p_e = 0.7 \cdot \min\left(1, \frac{e}{20}\right)
\end{equation}

This function increases linearly from 0 at the beginning of training to a ceiling of 0.7 by epoch 20, and remains constant thereafter. For each sample in a batch, we draw a Bernoulli variable $k \sim \mathrm{Bernoulli}(p_e)$ and:

\begin{itemize}
  \item If $k = 0$, the sample is left unchanged (no masking).
  \item If $k = 1$, we uniformly choose a single component index $c \in \{1, \dots, C\}$ and construct a binary mask $\mathcal{M} \in \{0,1\}^C$ such that $\mathcal{M}_c = 1$ and $\mathcal{M}_{c'} = 0$ for all $c' \ne c$. The masked input is then defined as:
  \begin{equation}
  \tilde{\mathbf{x}} = \mathbf{x} \odot \mathcal{M}
  \end{equation}
  where all but one component are zeroed out.
\end{itemize}

This curriculum strategy allows the model to first learn reliable reconstructions of complete architectures (with low masking probability), and then gradually adapt to the more challenging task of conditional generation as training progresses. During inference for conditional generation, we apply the appropriate mask to isolate the conditioning component.

\subsection{Masking VAE versus MMVAE\textsuperscript{+} in Component-Conditioned Generation}
\label{app:vae_vs_mmvae}

The unexpected outperformance of VAE over MMVAE\textsuperscript{+} can arise from four factors.

\paragraph{Information flow at inference}  
MMVAE\textsuperscript{+} infers the shared latent \(z\) from the available component’s encoder while the missing modalities receive non-informative priors for their private codes. Fine-grained spatial cues not captured by that single encoder are therefore lost.

\paragraph{Early feature sharing}  
Separate encoders prevent MMVAE\textsuperscript{+} from exposing convolutional filters to global part-to-part context in the first layers. The VAE’s single six-channel input allows joint spatial features to be learned throughout the network.

\paragraph{Masking curriculum}  
During training, the VAE randomly hides five channels and reconstructs the full architecture, directly conditioning the model to “fill in the gaps.” MMVAE\textsuperscript{+} receives no analogous pixel-level masking.

\paragraph{Hyper-parameter sensitivity}  
As reported in the original MMVAE\textsuperscript{+} paper, generative coherence degrades when the capacity of modality-specific subspaces is not tuned precisely; excessive private capacity bypasses the shared space and weakens connectivity priors.

Together, these factors might explain why the simpler, masking-trained VAE preserves single-region connectivity more reliably than the more complex MMVAE\textsuperscript{+} under the component-conditioned setting.

\subsection{Proof of proposition \ref{prop:masked_cat}}
\begin{proof}
Because each $r_k\ge 0$ and at least one $r_k \theta_k$ is strictly
positive, the normalising constant
$Z \;=\;\sum_{j} r_j \theta_j$ is finite and non-zero.
Define $\tilde{\theta}_k = r_k\theta_k / Z$.
\smallskip

\noindent\textbf{Non-negativity.}\ 
Since $r_k,\theta_k\ge 0$, each $\tilde{\theta}_k\ge 0$.

\noindent\textbf{Normalisation.}\ 
\[
\sum_{k=1}^{K} \tilde{\theta}_k
\;=\;
\frac{1}{Z}\sum_{k=1}^{K} r_k\theta_k
\;=\;1.
\]

\noindent Hence $\tilde{\boldsymbol{\theta}}$ is a valid categorical
parameter vector and the stated equality holds.
\end{proof}

\subsection{Architectures and Hyperparameters}
 Due to training instabilities, GANs were trained for up to 1000 epochs with checkpointing to retain the best-performing model.

\subsection{Component Descriptions}
\begin{itemize}    \item \textbf{\textit{Tread}}  Union of the tread
    rubber (\emph{bande de roulement}) and the steel or
    textile belt plies lying directly underneath.  This assembly fixes the
    outer diameter, footprint width and in-plane bending rigidity; it is
    therefore a very informative component for reconstructing the global
    envelope.    \item \textbf{\textit{Carcass}}  Aggregation of all
    carcass plies (\emph{nappes carcasse}) that wrap from
    bead to bead.  Together with the crown package it closes the toroidal
    cavity and controls inner air volume, making it a strong
    geometric anchor.\item \textbf{\textit{Sidewall}.}  External rubber
    (\emph{flanc}) that protects the carcass and allows flexion.  Its contour
    largely follows the two previous packages and thus provides moderate
    positional information.

    \item \textbf{\textit{Inner  Liner}.}  Thin, nearly texture-free
    but airtight rubber (\emph{gomme intérieure}) sealing the cavity.
    Because it mirrors the carcass shape while carrying little contrast, it is
    a weak standalone cue.    \item \textbf{\textit{Bead Filler}.}  Wedge-shaped rubber insert
    (\emph{bourrage de tringle}) that anchors the carcass to the bead wires and
    transmits forces to the rim.  Its footprint is small and confined to the
    lower sidewall.    \item \textbf{\textit{BAC} – Bead-Apex Cushion.}  Triangular pad just
    above the bead wires, filling the gap between \textit{BF} and the inner
    liner.  Like \textit{BF}, it covers only a few pixels in most slices.
\end{itemize}

\subsection{Model Architectures}
\label{app:architectures}

All five networks are built from the same ResNet building block shown in Table \ref{tab:resnet_block}.
Exactly how this block is arranged—how many times it is repeated, and where down-/up-sampling or attention is inserted—depends on the task, but the convolutional kernel, activation function and optional learned shortcut are identical across models.

\paragraph{Encoder / decoder backbone (GAN, $\beta$-VAE, MMVAE\textsuperscript{+}).}
For the three likelihood-based models we follow the generic encoder/decoder architecture in Tables \ref{tab:encoder_architecture} (down-path) and \ref{tab:decoder_architecture} (up-path).

\begin{itemize}
\item \textbf{GAN.}  The generator uses $k=4$ up-blocks with a base width of $nf=128$ and a cap of $nf_{\text{max}}=2048$; the latent prior is a 48-dimensional isotropic Gaussian.  The discriminator mirrors this depth, employs the same $nf$ schedule, and adds dropout ($p=0.1$) after every residual block.
\item \textbf{VAE.} Following the $\beta$‑VAE formulation \cite{higgins2017beta}, we performed a grid‑search over $\beta\in \{0.01,0.1,1,2.5,5\}$ and latent dimensions $d_z\in \{32,48,64\}$. The best configuration uses $\beta=2.5$, the same encoder/decoder depth as the GAN, a narrower base width ($nf=72$,$nf_{\text{max}}=1024$), and $d_z=48$.

\item \textbf{MMVAE\textsuperscript{+}.}  Six parallel encoders—one per modality—adopt the backbone of Table \ref{tab:encoder_architecture} with $k=3$ blocks, $nf=40$ and $nf_{\text{max}}=128$.  Each image is $(1{\times}64{\times}64)$, so the encoders see one channel each.  A seventh \emph{shared} encoder of identical shape produces the joint latent representation ($d_{z}=8$); modality-specific paths output $d_{w}=8$ per branch.  All decoders share parameters and follow Table \ref{tab:decoder_architecture}.
\end{itemize}

\paragraph{UNet backbone (DDPM, MDM).}
The diffusion models rely on the UNet template in Table \ref{tab:unet_architecture}.
Both use the same channel progression $\texttt{dim\_mults}=(1,2,4)$ with a base width $d_{0}=64$.

\begin{itemize}
\item \textbf{DDPM.}  The input consists of the six modalities concatenated along the channel axis, so $c_{\text{in}}=c_{\text{out}}=6$.  We enable flash attention in every resolution block.  Training follows the \textit{pred-$v$} objective with 1,000 diffusion steps.
\item \textbf{MDM.}  Before entering the UNet, the six modality images are collapsed into a single categorical map whose value range indexes $K+1$ classes.  The tensor therefore has shape $(n,1,64,64)$.  An embedding layer expands this to $64$ channels before the first convolution.  Apart from the categorical embedding and the fact that the final $1{\times}1$ convolution predicts $K+1$ logits per pixel, the architecture is identical to the DDPM UNet (same $\texttt{dim\_mults}$, attention placement, and time-embedding scheme).
\end{itemize}
\paragraph{Classifier free guidance.}
During inference we vary the guidance weight $w\in\{1,\dots,10\}$. 
Empirically, $w=4$ offers the best compromise: it minimises perceptual error (FID)  while maximising compliance with the prescribed 
dimensional constraints (DimErr) in the in‑distribution setting.


\begin{table*}[h]
\centering

\begin{tabular}{lccccc}
\toprule
\textbf{Layer} & \textbf{Input Chan.} & \textbf{Output Chan.} & \textbf{Kernel} & \textbf{Stride} & \textbf{Note}\\
\midrule
Conv2d (conv\_0) & $f_{\text{in}}$ & $f_{\text{hidden}}$ & $3{\times}3$ & 1 & Padding=1\\
Activation & -- & -- & -- & -- & LeakyReLU ($\alpha{=}0.2$)\\
optional Dropout & -- & -- & -- & -- & $p$ configurable\\
Conv2d (conv\_1) & $f_{\text{hidden}}$ & $f_{\text{out}}$ & $3{\times}3$ & 1 & Padding=1, bias on\\
Activation & -- & -- & -- & -- & LeakyReLU ($\alpha{=}0.2$)\\
\midrule
\textit{Shortcut} ($f_{\text{in}}\neq f_{\text{out}}$) & $f_{\text{in}}$ & $f_{\text{out}}$ & $1{\times}1$ & 1 & Learned, no padding\\
\bottomrule
\end{tabular}
\caption{Generic ResNet block used as the fundamental building block across all models}
\label{tab:resnet_block}
\end{table*}

\begin{table*}[h]

\centering
\begin{tabular}{llcl}
\toprule
\textbf{Stage} & \textbf{Layer(s)} & \textbf{Spatial Size} & \textbf{Comment} \\
\midrule
Input & $\mathrm{Image}$, shape $n\,{\times}\,c_{\text{in}}{\times}64{\times}64$ & $64{\times}64$ & $c_{\text{in}}\!=\!1$ (MMVAE$^+$) or $6$ (others)\\
Stem & Conv2d $(c_{\text{in}}\!\rightarrow\!nf,\;3{\times}3)$ & $64{\times}64$ & padding 1\\
\midrule
\multicolumn{4}{l}{\textit{Repeat $k$ times:}}\\
& AvgPool2d $(3{\times}3,\text{stride}=2)$ & $\bigl\lfloor\frac{s}{2}\bigr\rfloor$ & downsample\\
& ResNet Block & same &   \\
\midrule
Flatten & -- & $nf_k\cdot s_k^2$ & $s_k=64/2^{k}$ \\
Head & Fully-connected(s) & latent or $\mathbb{R}^1$ & $\mu$, $\log\sigma^2$ or real/fake score\\
\bottomrule
\end{tabular}
\caption{Generic encoder/discriminator architecture. Set $k{=}3$ for GAN/VAE, $k{=}3$ per-encoder in MMVAE$^+$, etc.}
\label{tab:encoder_architecture}
\end{table*}

\begin{table*}[h]
\centering
\begin{tabular}{llcl}
\toprule
\textbf{Stage} & \textbf{Layer(s)} & \textbf{Spatial Size} & \textbf{Comment} \\
\midrule
Latent FC & Linear $(d_{\!z}\!\rightarrow\!nf_0\cdot s_0^2)$ & $s_0{\times}s_0$ & $s_0=8$ \\
Reshape & View $\rightarrow (n,nf_0,s_0,s_0)$ & $8{\times}8$ & --\\
\midrule
\multicolumn{4}{l}{\textit{Repeat $k$ times (reverse order):}}\\
& ResNet Block & same &  \\
& Upsample $\times2$ & $2s$ & nearest or transposed conv\\
\midrule
Tail & ResNet Block & $64{\times}64$ & nf$\rightarrow$nf \\
Output Conv & Conv2d $(nf\!\rightarrow\!c_{\text{out}},3{\times}3)$ & $64{\times}64$ & $c_{\text{out}}\!=\!1$ (MMVAE$^+$) or $6$ (others)\\
\bottomrule
\end{tabular}
\caption{Generic decoder/generator architecture. Choose $k$ to mirror the encoder depth from Table \ref{tab:encoder_architecture}.}\label{tab:decoder_architecture}
\end{table*}

\begin{table*}[h]
\centering
\begin{tabular}{llcl}
\toprule
\textbf{Phase} & \textbf{Block(s)} & \textbf{Channels} & \textbf{Notes} \\
\midrule
Input Stem & Conv2d $(c_{\text{in}}\!\rightarrow\!d_0,\;7{\times}7)$ & $d_0$ & padding 3 \\
\midrule
\multicolumn{4}{l}{\textbf{Down path} (for $\ell=0\dots L{-}1$)}\\
& \quad ResNet Block $\times2$ & $d_\ell$ & time-embedding injection\\
& \quad Attention (Linear or Full) & $d_\ell$ & optional\\
& \quad Downsample ($2{\times}$) & $d_{\ell+1}$ & conv-stride 2 or space-to-depth\\
\midrule
Bottleneck & ResNet Block -- Attention -- ResNet Block & $d_L$ & --\\
\midrule
\multicolumn{4}{l}{\textbf{Up path} (mirror $\ell=L{-}1\dots0$)}\\
& \quad Concat skip + ResNet Block $\times2$ & $d_\ell$ &  \\
& \quad Attention & $d_\ell$ & optional\\
& \quad Upsample ($\times2$) & $d_{\ell-1}$ & nearest + conv or transposed conv\\
\midrule
Final Head & ResNet Block → Conv2d $(d_0\!\rightarrow\!c_{\text{out}},1{\times}1)$ & $c_{\text{out}}$ & logits/noise prediction \\
\bottomrule
\end{tabular}
\caption{UNet architecture used by DDPM and MDM models. Depth $L$ corresponds to \texttt{dim\_mults} length (e.g., $(1,2,4)\Rightarrow\ L=3$). Attention type and dropout match implementation.}
\label{tab:unet_architecture}
\end{table*}

\end{document}